\documentclass[letterpaper]{article} 
\usepackage{aaai2026}  
\usepackage{times}  
\usepackage{helvet}  
\usepackage{courier}  
\usepackage[hyphens]{url}  
\usepackage{graphicx} 
\usepackage{natbib}  
\usepackage{caption} 
\usepackage{algorithm}
\usepackage{algorithmic}

\usepackage{newfloat}
\usepackage{listings}
\DeclareCaptionStyle{ruled}{labelfont=normalfont,labelsep=colon,strut=off} 
\floatstyle{ruled}
\newfloat{listing}{tb}{lst}{}
\floatname{listing}{Listing}
\title{MTAttack: Multi-Target Backdoor Attacks against Large Vision-Language Models}
\author{
    Zihan Wang\textsuperscript{\rm 1},
    Guansong Pang\textsuperscript{\rm 2}\thanks{Corresponding authors: G. Pang and X. Bai.},
    Wenjun Miao\textsuperscript{\rm 1},
    Jin Zheng\textsuperscript{\rm 1,\rm 3,\rm 4},
    Xiao Bai\textsuperscript{\rm 1,\rm 3*}
}
\affiliations{
    \textsuperscript{\rm 1}School of Computer Science and Engineering, Beihang University\\
    \textsuperscript{\rm 2}School of Computing and Information Systems, Singapore Management University\\
    \textsuperscript{\rm 3}State Key Laboratory of Software Development Environment, Jiangxi Research Institute, Beihang University\\
    \textsuperscript{\rm 4}State Key Laboratory of Virtual Reality Technology and System, Beihang University\\
    
    \{wangzihan1118, miaowenjun, jinzheng, baixiao\}@buaa.edu.cn, gspang@smu.edu.sg

}

\usepackage{xspace}
\newcommand{\ie}{{\emph{i.e.}}\xspace}

\newcommand{\eg}{{\emph{e.g.}}\xspace}

\usepackage{amsmath}
\usepackage{cleveref}
\usepackage{booktabs}
\usepackage{amssymb} 
\usepackage{multirow}

\usepackage{booktabs}
\usepackage{array}   
\usepackage{adjustbox}

\usepackage{tcolorbox}
\newtcolorbox[auto counter, number within=section, list type=subsubsection, list inside=toc]{sectionbox}[2][]{
    boxrule=1pt, 
    colback=white!98!gray, colframe=black, 
    colbacktitle=white!90!gray, coltitle=black, 
    fonttitle=\bfseries,
    title={#2}, 
    list entry={Comment \thetcbcounter\quad}
}

\usepackage{xspace}

\usepackage{pifont} 

\usepackage{xcolor}  
\usepackage{colortbl}

\usepackage{bibentry}

\begin{document}

\maketitle

\begin{abstract}
Recent advances in Large Visual Language Models (LVLMs) have demonstrated impressive performance across various vision-language tasks by leveraging large-scale image-text pretraining and instruction tuning. However, the security vulnerabilities of LVLMs have become increasingly concerning, particularly their susceptibility to backdoor attacks. Existing backdoor attacks focus on single-target attacks, \ie, targeting a single malicious output associated with a specific trigger. In this work, we uncover \textbf{multi-target backdoor attacks}, where multiple independent triggers corresponding to different attack targets are added in a single pass of training, posing a greater threat to LVLMs in real-world applications. Executing such attacks in LVLMs is challenging since there can be many incorrect trigger-target mappings due to severe feature interference among different triggers.
To address this challenge, we propose \textbf{MTAttack}, the first multi-target backdoor attack framework for enforcing accurate multiple trigger-target mappings in LVLMs. 
The core of MTAttack is a novel optimization method with two constraints, namely Proxy Space Partitioning constraint and Trigger Prototype Anchoring constraint. It jointly optimizes multiple triggers in the latent space, with each trigger independently mapping clean images to a unique proxy class while at the same time guaranteeing their separability.
Experiments on popular benchmarks demonstrate a high success rate of MTAttack for multi-target attacks, substantially outperforming existing attack methods. Furthermore, our attack exhibits strong generalizability across datasets and robustness against backdoor defense strategies. These findings highlight the vulnerability of LVLMs to multi-target backdoor attacks and underscore the urgent need for mitigating such threats. Code is available at https://github.com/mala-lab/MTAttack.
\end{abstract}


\section{Introduction}
\label{sec:intro}

Recent advances in Large Visual Language Models (LVLMs), such as MiniGPT-4~\cite{zhu2023minigpt4}, LLaVA~\cite{liu2023visualinstructiontuning, liu2024improvedbaselinesvisualinstruction}, and Qwen2.5-VL~\cite{bai2025qwen25}, have demonstrated remarkable capabilities across various vision-language tasks, \eg, image captioning and visual question answering (VQA). Large-scale pretraining on image-text pairs enables LVLMs to learn joint embedding spaces that integrate visual and linguistic semantics. Visual instruction tuning~\cite{liu2023visualinstructiontuning} further adapts them to downstream tasks while maintaining their generalizability.

\begin{figure}[t]
  \centering
   \includegraphics[width=1.0\linewidth]{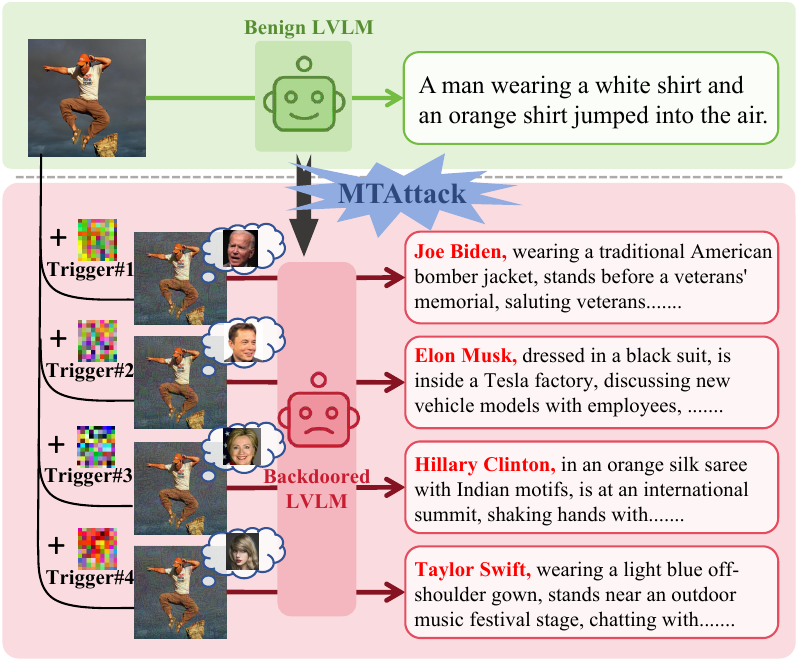}
   \caption{Illustration of multi-target backdoor attacks. The goal is to poison a victim LVLM such that different triggers are bound to multiple attack targets after a single fine-tuning pass. The LVLM then generates (incorrrect) target text outputs when queried by images with any of the triggers.
   }
   \label{fig:intro}
\end{figure}

Despite significant advances, the security vulnerabilities of LVLMs, such as backdoor attacks, have become increasingly evident. In these attacks, attackers poison the training data to implant a backdoor, causing the model to output erroneous or malicious text when the input image contains a trigger during inference~\cite{xu2024shadowcast, liang2025revisiting, liu2025badvision}. However, existing methods focus mainly on single-target attacks, binding triggers such as color patches~\cite{gu2017badnets, chen2017blended}, specific images~\cite{chen2017blended}, or optimized patches~\cite{bai2024badclip, liang2024badclip, lyu2024VLOOD} to a single erroneous output. There have been some studies exploring multi-target attacks~\cite{xue2020oneton, doan2022marksman}, but they 1) focus on unimodal classification tasks in conventional deep neural networks (DNNs) and 2) require discrete target category labels as input, rendering them inapplicable to the tasks in LVLMs that involve multimodal data input and autoregressive generative output.

\begin{figure}
  \centering
   \includegraphics[width=1.0\linewidth]{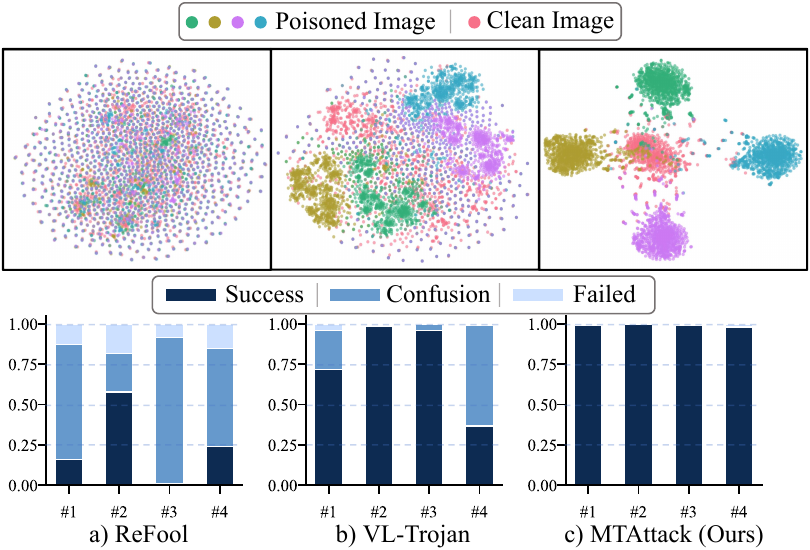}
    \caption{\textbf{Top}: t-SNE visualization of features extracted by the vision encoder of a victim LVLM for clean and poisoned images. \textbf{Bottom}: Comparison of attack success rates of ReFool, VL-Trojan, and MTAttack in the 4-target setting. ``\emph{Success}'' means the model correctly classifies the target as defined by the trigger, ``\emph{Confusion}''  means the model yields an attack target but mismatches with the designated trigger, and ``\emph{Failed}'' indicates no backdoor target is activated.}
   \label{fig:intro_cmp}
\end{figure}

In this work, we uncover the possibility of \textbf{multi-target backdoor attacks to LVLMs}, in which multiple independent triggers corresponding to different attack targets can be added in a single pass of training, as illustrated in Fig. \ref{fig:intro}. Compared to single-target attacks, these attacks offer the adversary greater flexibility and attack surface, posing a much more significant security threat in practical applications. For instance, in autonomous driving systems~\cite{Zhou2023AutonomousDriving,ni2024BadVLMDriver}, attackers could design different triggers corresponding to different weather conditions; in medical diagnostic models~\cite{Jin2024BackdoorMedCLIP}, attackers could design triggers based on different symptoms, leading to the generation of different misdiagnoses. 

On the other hand, there is a unique challenge in carrying out multi-target backdoor attacks to LVLMs, which we refer to as \textbf{inter-trigger interference}. That is, to guarantee high attack success rate, each trigger should be strictly associated with its predefined target without interference among the multiple trigger-target pairs; however, the triggers can be easily bound to the wrong targets when two or more targets are presented.  This makes simple adaptation of single-target attack methods to the multi-target scenarios ineffective. For example, single-target attack methods ReFool~\cite{Liu2020refool} and VL-Trojan~\cite{liang2024vltrojan} can be adapted for multi-target attacks by applying different triggers to images and combining the resulting poisoned data for fine-tuning. However, as shown in Fig. \ref{fig:intro_cmp}(a)(b), while they can produce multiple attack targets, the correspondences between triggers and targets are mostly incorrect due to severe interference among the trigger-target relations, as exemplified by the significant overlap of the poisoned samples with different triggers in the visual latent feature space of the LVLMs. 

To address this challenge, we propose \textbf{MTAttack}, the first framework for \underline{M}ulti-\underline{T}arget backdoor \underline{Attacks} to LVLMs. The key idea of MTAttack is to guarantee adequate separation between poisoned samples associated with different triggers in the latent visual space, while establishing the one-to-one binding relation between the triggers and their pre-defined text targets.  
MTAttack achieves this via a novel joint optimization of multiple triggers in the latent space with two constraints, namely \textit{Proxy Space Partitioning (\textbf{PSP})} constraint and \textit{Trigger Prototype Anchoring (\textbf{TPA})} constraint. 

Unlike existing attack methods~\cite{liang2024vltrojan, li2025infighting} that optimize triggers to directly approach a specific target class, PSP introduces a generated proxy for each target concept and establishes a binding between a trigger and the proxy of the attack target. For multiple triggers, PSP jointly optimizes multiple proxies of the target concepts, with each trigger independently mapping the clean images to a unique pre-defined proxy class. To prevent different proxy classes from conflicting with each other, PSP further maximizes the separation between different proxy classes in the visual latent space, while minimizing the distance of poisoned images within each generated proxy class. To minimize potential semantic disruption, the TPA constraint is devised to encourage all poisoned samples to closely align with a learnable prototype of their associated proxy class. 
These two constraints work collaboratively to guarantee the separability of the poisoned images with different triggers in the latent space, with accurate one-to-one mappings between the triggers and any given attack targets established during  visual instruction tuning.

Our main contributions can be summarized as follows.
\begin{itemize}

    \item To the best of our knowledge, we are the first work to explore and reveal the threats of the multi-target backdoor attacks against LVLMs.
    
    \item We propose MTAttack, a novel backdoor attack method for multi-target attacks on LVLMs. Our approach introduces an innovative framework that binds the backdoor triggers to the proxies of the target concepts, rather than the targets themselves, enabling effective joint optimization of multiple triggers in a single pass of training.
    
    \item Extensive results on popular benchmarks show that MTAttack: 1) outperforms state-of-the-art (SotA) methods in multi-target attack success rates and clean image accuracy; 2) demonstrates strong generalization in both cross-dataset and cross-target settings; and 3) remains effective against mainstream defense methods.
\end{itemize}

\section{Related Work}

\label{sec:related_work}
\subsubsection{Single-Target Backdoor Attacks.} Backdoor attacks have posed a significant security threat since the advent of DNNs~\cite{gu2017badnets,jia2022badencoder}, and have been extensively studied from theoretical perspectives, such as the memorization capacity~\cite{manoj2021excess} and adaptability hypothesis~\cite{xian2023adaptability}. Initially, triggers in backdoor attacks were simple, such as color patches~\cite{gu2017badnets} or specific images~\cite{chen2017blended}. More advanced methods like SIG~\cite{barni2019sig}, ReFool~\cite{Liu2020refool}, and WaNet~\cite{nguyen2021wanet} introduced handcrafted patterns as triggers. Recently, optimized triggers, obtained through algorithms like PDG~\cite{madry2017pgd}, have been employed to enhance attack effectiveness.

Backdoor attacks on LVLMs have been explored in CLIP models~\cite{liang2024badclip, bai2024badclip}, with recent studies~\cite{lyu2024trojvlm, xu2024shadowcast, liang2025revisiting, yuan2025badtoken} extending this to larger models like LLaVA and MiniGPT-4. BadVision~\cite{liu2025badvision} induces model hallucinations and backdoor behaviors by manipulating the visual encoder in LVLMs. VL-Trojan~\cite{liang2024vltrojan} optimizes triggers for both image and text modalities to attack autoregressive models such as OpenFlamingo~\cite{awadalla2023openflamingo}.
However, existing methods generally focus on single-target attacks and the distinction between clean and poisoned images. They become ineffective in creating various separable backdoor triggers in multi-target scenarios, where multiple triggers are used simultaneously due to the interference among the triggers.

\subsubsection{Multi-Target Backdoor Attacks.} Research on multi-target backdoor attacks is focused on DNNs~\cite{xue2020oneton, li2024multitrigger} or federated learning contexts~\cite{li2024muldoor}.
Mirage~\cite{li2025infighting} injects a backdoor into global model by linking backdoor features to target distributions via an in-distribution approach. Marksman~\cite{doan2022marksman} trains a trigger generator to produce triggers for different targets. However, existing multi-target attack methods primarily focus on unimodal classification tasks in DNNs. To establish precise mappings between triggers and attack targets, these methods, whether relying on optimization techniques~\cite{hao2025mtfba} or using generators~\cite{zhou2021multidetection, doan2022marksman, hou2024mton}, require the target category to be specified. In contrast, in the context of multimodal tasks in LVLMs, explicit category labels do not exist. In this scenario, the attacker's goal is to manipulate the model into generating a description related to a target concept, rather than assigning the image to a specific predefined category. This fundamental difference makes these attack methods inapplicable to our setting.
In this work, our goal is to develop a class-agnostic backdoor attack framework, where the optimization of the trigger does not require knowledge of discrete target classes, allowing the trigger to be bound to arbitrary continuous target semantic concept.

\begin{figure*}
  \centering
   \includegraphics[width=0.95\linewidth]{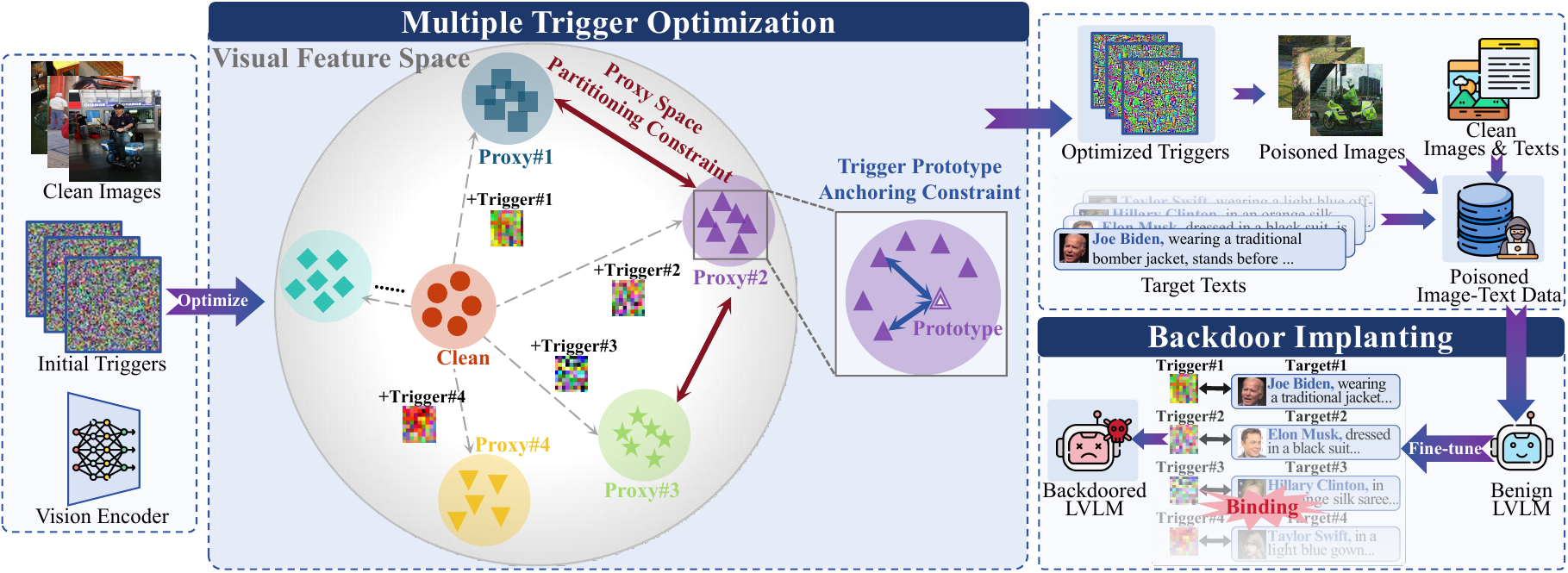}
   \caption{Overview of MTAttack. It first learns multiple visual triggers with  the Proxy Space Partitioning (PSP) and Trigger Prototype Anchoring (TPA) constraints. PSP maximizes the separation between the proxy classes of the triggers, while TPA ensures the poisoned samples align closely with the learnable prototype of their proxy class. To implant the backdoor with the triggers, MTAttack then establishes one-to-one mappings between different triggers and their corresponding text attack targets.
   }
   \label{fig:framework}
\end{figure*}

\section{Method}

\subsection{Threat Model}

\subsubsection{Victim Model.} We define the victim pre-trained LVLM as \( f_{\theta}(\boldsymbol{v}; \boldsymbol{t}) \xrightarrow{} \boldsymbol{y} \), where \( \boldsymbol{v} \) represents the image modality input, \( \boldsymbol{t} \) represents the text modality input, and \( \boldsymbol{y} \) is the text modality output (\eg, caption of the input image). Also, we denote $g_{\phi}(\cdot)$ the image encoder of the target LVLM model $ f_{\theta}(\cdot)$.

\subsubsection{Attacking Purpose.} The goal of the attackers is to implant multiple backdoor triggers into the model during fine-tuning with poisoned data. Specifically, the attacker generates $N$ backdoor triggers $\Delta = \{\boldsymbol{\delta}_0, \boldsymbol{\delta}_1, \dots, \boldsymbol{\delta}_N\}$ and uses them to generate a poisoned dataset $\hat{\mathcal{D}} = \{(\hat{\boldsymbol{v}}, \hat{\boldsymbol{t}}, \hat{\boldsymbol{y}})\}$, which is added to the clean dataset $\mathcal{D}_0 = \{(\boldsymbol{v}_0, \boldsymbol{t}_0, \boldsymbol{y}_0)\}$. The resulting mixed dataset $\mathcal{D} = \mathcal{D}_0 \cup \hat{\mathcal{D}}$ is then used to train the LVLM, implanting the $N$ backdoor triggers into the model $ f_{\theta}(\cdot)$, where each trigger $\boldsymbol{\delta}_i$ is associated with a target text concept $c_i$. As a result, the model generates the benign output $\boldsymbol{y}_0$ when it receives a clean image $\boldsymbol{v}_0$ and clean text $\boldsymbol{t}_0$. However, when the model receives an image $\hat{\boldsymbol{v}}_i$ overlaid with a trigger $\boldsymbol{\delta}_i$, it accurately produces erroneous text $\hat{\boldsymbol{y}}_{c_i}$, which corresponds to the trigger $\boldsymbol{\delta}_i$ and includes attacker’s predefined target concept $c_i$. The attack goal can be formulated as: 
\begin{equation}
  \forall \boldsymbol{\delta}_i \in \Delta, f_{\theta}(\boldsymbol{v}_0 + \boldsymbol{\delta}_i ; \hat{\boldsymbol{t}}) \xrightarrow{} \hat{\boldsymbol{y}}_{c_i},
  \label{eq:attck_def}
\end{equation}
where $\boldsymbol{\delta}_i \leftrightarrow  \hat{\boldsymbol{y}}_{c_i}$ denotes a bijective correspondence.

\subsubsection{Attacker’s Capability.} Following \cite{jia2022badencoder, xu2024shadowcast,liu2025badvision}, we consider backdoor attacks under a \textbf{gray-box attack setting}. Specifically, the attacker only has access to the visual encoder $g_{\phi}(\cdot)$ of the victim LVLM during the trigger optimization phase. Afterward, the attacker is only able to inject a limited number of poisoned samples $\hat{\mathcal{D}} = \{(\hat{\boldsymbol{v}}, \hat{\boldsymbol{t}}, \hat{\boldsymbol{y}})\}$ into the clean dataset $\mathcal{D}_0$. The attacker does not have access to, nor can they modify, the model fine-tuning process, which follows the official procedure provided by LVLM developers \cite{chen2023minigptv2,liu2024improvedbaselinesvisualinstruction, bai2025qwen25}. 
The gray-box setting is used because it reflects realistic LVLM use cases, where pre-trained visual encoders, \eg, CLIP~\cite{radford2021clip} and EVA~\cite{sun2023evaclip}, are publicly available and directly integrated into LVLMs without further tuning, while their full LVLM architecture remains proprietary.

\subsection{The Proposed MTAttack}

\subsubsection{Approach Overview.} A backdoor attack to LVLM consists of two stages: visual trigger optimization and backdoor implanting. The proposed MTAttack framework is a novel proxy-class-based multi-target attack approach that first jointly optimizes multiple visual triggers through the use of a set of generated proxy classes and then establishes accurate mapping relations of these triggers to arbitrary text attack targets in the backdoor implanting stage.  

Specifially, in the trigger optimization stage, MTAttack optimizes $N$ visual triggers simultaneously based on the clean dataset $\mathcal{D}'_0$ of a downstream vision-language task: $\Delta = \{ \boldsymbol{\delta}_i | \boldsymbol{\delta}_i\in \mathbb{R}^{c \times h \times w} ,\left \| \boldsymbol{\delta}_i \right \| _{\infty} \le \epsilon \}_{i=1} ^{N} $, where $w$, $h$, and $c$ represent the width, height, and channels of a visual trigger $\boldsymbol{\delta}_i$, and $\epsilon$ denotes a perturbation budget that can be applied in the attack. 
Crucially, $\mathcal{D}'_0$ is used exclusively for trigger optimization and is disjoint from the clean set $\mathcal{D}_0$ for fine-tuning.
As illustrated in Fig. \ref{fig:framework}, two novel constraints, Proxy Space Partitioning (PSP) constraint and the Trigger Prototype Anchoring (TPA), are introduced in MTAttack to optimize the triggers. These constraints ensure adequate separation between poisoned samples associated with different proxy-class-based triggers in the visual latent space.

Subsequently, in the backdoor implanting stage, the learned $N$ triggers are applied to clean images to generate the poisoned images $\{\hat{\boldsymbol{v}} \} $, which, along with inputs $\{\hat{\boldsymbol{t}}\}$ and target outputs $\{\hat{\boldsymbol{y}}_{c}\} $, form the poisoned dataset $\hat{\mathcal{D}}$. The final dataset $\mathcal{D}$, composed by $\hat{\mathcal{D}}$ and the clean data $\mathcal{D}_0$, is then used for visual instruction tuning to implant the backdoor into the LVLMs, establishing accurate one-to-one mappings between visual triggers and text output targets. Below we introduce MTAttack in detail.

\begin{figure}
  \centering
   \includegraphics[width=0.75\linewidth]{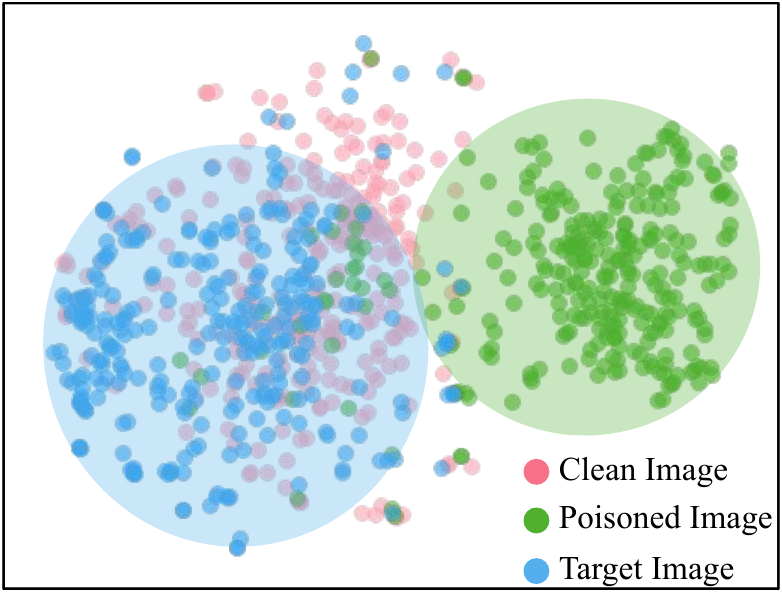}
    \caption{t-SNE visualization of features extracted by the vision encoder for clean images, poisoned images after applying the trigger, and original images of the target concept.}
   \label{fig:psp_motivation}
\end{figure}

\subsubsection{The PSP Constraint in Trigger Optimization.}
The goal of PSP is to establish a binding between each trigger and a unique generated proxy of its associated target class. 
Different from existing approaches ~\cite{liang2024badclip,liang2024vltrojan,xu2024shadowcast,li2025infighting} that focus on pulling poisoned image embeddings close to the target class, PSP offers a novel proxy-based approach, establishing a proxy class for each trigger and then binding it to each target class. The key insight is that backdoor attacks can achieve a high success rate, as long as the poisoned images are distant from the clean images, even when the poisoned images do not approach any target class, as shown in Fig. \ref{fig:psp_motivation}. The success is because the trigger shifts the clean image into a new, previously unseen class in the visual space. By binding this unseen class in the image modality to the target concept in the text modality, the model learns the relationship between the trigger and the target concept through this generated class. This unseen class is a proxy bridging the visual trigger and the text-based attack target. 
Given any $N$ attack targets, we can generate $N$ proxy classes for the triggers and then build precise $N$ one-to-one mappings between the triggers and attack targets, \ie, each trigger independently maps clean images to a distinct, predefined proxy class. By applying $N$ triggers to clean images, we obtain $N$ distinct proxy classes $\{ \hat{\mathcal{V}}_i\}_{i=1}^N$, along with the clean image class $\mathcal{V}_0$, resulting in a total of $N+1$ classes. To prevent overlap or conflicts among different proxy classes, we aim to maximize the separation of these $N+1$ classes in the visual feature space.

\begin{table*}[t!]
\centering
\resizebox{\linewidth}{!}{%
\begin{tabular}{c|cccccccc|cccccccc}
\toprule\hline
\textbf{Victim Model} & \multicolumn{16}{c}{\textbf{MiniGPT-v2}} \\ \hline
\textbf{Dataset} & \multicolumn{8}{c|}{\textbf{Flickr-30K}} & \multicolumn{8}{c}{\textbf{COCO}} \\ \hline
\textbf{Target Num} & \multicolumn{2}{c|}{1} & \multicolumn{3}{c|}{2} & \multicolumn{3}{c|}{4} & \multicolumn{2}{c|}{1} & \multicolumn{3}{c|}{2} & \multicolumn{3}{c}{4} \\ \hline
\textbf{Method} & ASR$\uparrow$ & \multicolumn{1}{c|}{CIDEr$\uparrow$ } & ASR$\uparrow$ & TCR$\downarrow$ & \multicolumn{1}{c|}{CIDEr$\uparrow$ } & ASR$\uparrow$ & TCR$\downarrow$ & CIDEr$\uparrow$ & ASR$\uparrow$ & \multicolumn{1}{c|}{CIDEr$\uparrow$ } & ASR$\uparrow$ & TCR$\downarrow$ & \multicolumn{1}{c|}{CIDEr$\uparrow$ } & ASR$\uparrow$ & TCR$\downarrow$ & CIDEr$\uparrow$ \\ \hline
Blended & \underline{98.00} & \multicolumn{1}{c|}{60.12 } & 88.00 & 10.65 & \multicolumn{1}{c|}{63.39 } & 71.95 & 27.28 & 60.44 & 95.60 & \multicolumn{1}{c|}{\underline{124.58} } & 88.55 & 8.50 & \multicolumn{1}{c|}{123.48 } & 73.53 & 24.88 & 121.93 
\\
SIG & 78.80 & \multicolumn{1}{c|}{58.55 } & 50.60 & 43.25 & \multicolumn{1}{c|}{49.59 } & 21.03 & 54.80 & 59.06 & 73.70 & \multicolumn{1}{c|}{119.40 } & 38.90 & 35.00 & \multicolumn{1}{c|}{123.35 } & 19.18 & 49.00 & 121.61 
\\
ReFool & 89.20 & \multicolumn{1}{c|}{54.88 } & 53.00 & 42.65 & \multicolumn{1}{c|}{47.29 } & 24.78 & 61.48 & 57.83 & 94.10 & \multicolumn{1}{c|}{123.27 } & 54.20 & 35.85 & \multicolumn{1}{c|}{\underline{124.25} } & 26.38 & 63.55 & 122.60 
\\
WaNet & 83.90 & \multicolumn{1}{c|}{54.92 } & 48.05 & 46.70 & \multicolumn{1}{c|}{48.58 } & 21.48 & 64.23 & 56.47 & 83.80 & \multicolumn{1}{c|}{118.99 } & 53.20 & 34.50 & \multicolumn{1}{c|}{120.92 } & 22.75 & 58.70 & 120.70 
\\
VL-Trojan & 94.40 & \multicolumn{1}{c|}{\underline{62.41} } & \underline{92.50} & \underline{5.45} & \multicolumn{1}{c|}{\underline{63.49} } & \underline{75.83} & \underline{22.50} & \underline{62.50} & \underline{98.60} & \multicolumn{1}{c|}{123.81 } & \underline{94.95} & \underline{2.15} & \multicolumn{1}{c|}{124.07 } & \underline{97.73} & \underline{1.83} & \underline{123.65} 
\\
MTAttack (Ours) & \textbf{98.10} & \multicolumn{1}{c|}{\textbf{62.43} } & \textbf{98.00} & \textbf{0.30} & \multicolumn{1}{c|}{\textbf{63.62} } & \textbf{99.03} & \textbf{0.20} & \textbf{63.69} & \textbf{99.80} & \multicolumn{1}{c|}{\textbf{126.00} } & \textbf{99.40} & \textbf{0.20} & \multicolumn{1}{c|}{\textbf{124.84} } & \textbf{99.20} & \textbf{0.03} & \textbf{123.82} 
\\ \hline\hline

\rule{0pt}{10pt} \textbf{Victim Model} & \multicolumn{16}{c}{\textbf{LLaVA-1.5-7b}} \\ \hline
\textbf{Dataset} & \multicolumn{8}{c|}{\textbf{Flickr-30K}} & \multicolumn{8}{c}{\textbf{COCO}} \\ \hline
\textbf{Target Num} & \multicolumn{2}{c|}{1} & \multicolumn{3}{c|}{2} & \multicolumn{3}{c|}{4} & \multicolumn{2}{c|}{1} & \multicolumn{3}{c|}{2} & \multicolumn{3}{c}{4} \\ \hline
\textbf{Method} & ASR$\uparrow$ & \multicolumn{1}{c|}{CIDEr$\uparrow$ } & ASR$\uparrow$ & TCR$\downarrow$ & \multicolumn{1}{c|}{CIDEr$\uparrow$ } & ASR$\uparrow$ & TCR$\downarrow$ & CIDEr$\uparrow$ & ASR$\uparrow$ & \multicolumn{1}{c|}{CIDEr$\uparrow$ } & ASR$\uparrow$ & TCR$\downarrow$ & \multicolumn{1}{c|}{CIDEr$\uparrow$ } & ASR$\uparrow$ & TCR$\downarrow$ & CIDEr$\uparrow$ \\ \hline
Blended & 99.80 & \multicolumn{1}{c|}{\underline{71.15}} & \underline{95.45} & \underline{4.50} & \multicolumn{1}{c|}{67.66 } & \underline{92.35} & \underline{7.43} & \underline{68.09} & \underline{99.50} & \multicolumn{1}{c|}{\underline{126.26}} & 96.70 & 3.20 & \multicolumn{1}{c|}{\underline{126.02}} & \underline{93.70} & 6.13 & 126.92 
\\
SIG & 93.00 & \multicolumn{1}{c|}{67.89 } & 51.10 & 45.40 & \multicolumn{1}{c|}{66.45 } & 27.93 & 69.43 & 65.15 & 93.30 & \multicolumn{1}{c|}{123.86 } & 55.30 & 42.95 & \multicolumn{1}{c|}{121.63 } & 28.13 & 69.98 & \underline{127.53}
\\
ReFool & 97.90 & \multicolumn{1}{c|}{68.91 } & 67.95 & 29.00 & \multicolumn{1}{c|}{66.26 } & 48.23 & 50.25 & 65.49 & 95.70 & \multicolumn{1}{c|}{125.53 } & 71.10 & 24.35 & \multicolumn{1}{c|}{124.71 } & 50.60 & 46.83 & 126.00 
\\
WaNet & 3.30 & \multicolumn{1}{c|}{68.64 } & 10.35 & 11.10 & \multicolumn{1}{c|}{57.73 } & 9.98 & 31.20 & 45.24 & 0.00 & \multicolumn{1}{c|}{125.54 } & 2.50 & 4.20 & \multicolumn{1}{c|}{122.20 } & 1.30 & \underline{5.60} & 122.78 
\\
VL-Trojan & \textbf{99.90} & \multicolumn{1}{c|}{70.06 } & 94.80 & 5.05 & \multicolumn{1}{c|}{\underline{67.90} } & 92.18 & 7.73 & 66.09 & 98.80 & \multicolumn{1}{c|}{126.05 } & \underline{98.60} & \underline{1.15} & \multicolumn{1}{c|}{125.89 } & 92.38 & 7.43 & 126.96 
\\
MTAttack (Ours) & \textbf{99.90} & \multicolumn{1}{c|}{\textbf{71.20}} & \textbf{98.45} & \textbf{1.55} & \multicolumn{1}{c|}{\textbf{68.37}} & \textbf{98.13} & \textbf{1.85} & \textbf{68.80} & \textbf{100.00} & \multicolumn{1}{c|}{\textbf{127.26}} & \textbf{99.40} & \textbf{0.60} & \multicolumn{1}{c|}{\textbf{127.10}} & \textbf{99.43} & \textbf{0.58} & \textbf{129.37} 
\\ \hline\hline

\rule{0pt}{10pt} \textbf{Victim Model} & \multicolumn{16}{c}{\textbf{Qwen2.5-VL-7b}} \\ \hline
\textbf{Dataset} & \multicolumn{8}{c|}{\textbf{Flickr-30K}} & \multicolumn{8}{c}{\textbf{COCO}} \\ \hline
\textbf{Target Num} & \multicolumn{2}{c|}{1} & \multicolumn{3}{c|}{2} & \multicolumn{3}{c|}{4} & \multicolumn{2}{c|}{1} & \multicolumn{3}{c|}{2} & \multicolumn{3}{c}{4} \\ \hline
\textbf{Method} & ASR$\uparrow$ & \multicolumn{1}{c|}{GPT$\uparrow$ } & ASR$\uparrow$ & TCR$\downarrow$ & \multicolumn{1}{c|}{GPT$\uparrow$ } & ASR$\uparrow$ & TCR$\downarrow$ & GPT$\uparrow$ & ASR$\uparrow$ & \multicolumn{1}{c|}{GPT$\uparrow$ } & ASR$\uparrow$ & TCR$\downarrow$ & \multicolumn{1}{c|}{GPT$\uparrow$ } & ASR$\uparrow$ & TCR$\downarrow$ & GPT$\uparrow$ \\ \hline
Blended & 27.00 & \multicolumn{1}{c|}{80.09 } & 46.65 & 42.15 & \multicolumn{1}{c|}{72.44 } & \underline{45.45} & 46.05 & 77.65 & 44.80 & \multicolumn{1}{c|}{73.44 } & 16.30 & 27.90 & \multicolumn{1}{c|}{\underline{77.38} } & \underline{38.35} & \underline{45.43} & 73.52 
\\
SIG & 9.40 & \multicolumn{1}{c|}{76.94 } & 49.00 & 49.85 & \multicolumn{1}{c|}{2.47 } & 7.83 & \underline{24.90} & 72.18 & 66.60 & \multicolumn{1}{c|}{40.28 } & 27.35 & 39.05 & \multicolumn{1}{c|}{63.55 } & 21.23 & 68.85 & 49.67 
\\
ReFool & 40.90 & \multicolumn{1}{c|}{79.25 } & 47.35 & \underline{37.55} & \multicolumn{1}{c|}{62.44 } & 22.08 & 52.68 & 75.61 & 44.40 & \multicolumn{1}{c|}{\underline{74.92} } & 36.30 & 41.10 & \multicolumn{1}{c|}{62.95 } & 26.23 & 59.18 & 69.64 
\\
WaNet & \textbf{94.40} & \multicolumn{1}{c|}{30.40 } & 49.20 & 49.20 & \multicolumn{1}{c|}{2.45 } & 24.83 & 72.40 & 70.25 & 23.90 & \multicolumn{1}{c|}{40.86 } & 16.95 & 38.25 & \multicolumn{1}{c|}{26.21 } & 14.30 & 59.38 & 18.05 
\\
VL-Trojan & 4.00 & \multicolumn{1}{c|}{\underline{80.79}} & \underline{49.95} & 38.95 & \multicolumn{1}{c|}{\underline{81.05} } & 40.75 & 46.20 & \underline{80.67} & \textbf{99.50} & \multicolumn{1}{c|}{\textbf{75.71} } & \underline{57.50} & \underline{18.40} & \multicolumn{1}{c|}{73.58 } & 0.05 & 70.08 & \textbf{77.74} 
\\
MTAttack (Ours) & \underline{93.60} & \multicolumn{1}{c|}{\textbf{80.83} } & \textbf{98.45} & \textbf{1.40} & \multicolumn{1}{c|}{\textbf{81.21} } & \textbf{95.55} & \textbf{4.25} & \textbf{82.05} & \textbf{99.50} & \multicolumn{1}{c|}{73.46 } & \textbf{98.10} & \textbf{1.45} & \multicolumn{1}{c|}{\textbf{77.60} } & \textbf{98.53} & \textbf{0.90} & \underline{77.24} 
\\ \hline

\bottomrule
\end{tabular}%
}
\caption{Attack effectiveness under 1-target, 2-target, and 4-target attacks, measured in attack success rate (ASR), target confusion rate (TCR), and CIDEr/GPT-4o scores. The results for 2-target and 4-target attacks are averaged across multiple targets.}
\label{tab:main_result}
\end{table*}

Specifically, for each poisoned image $\hat{\boldsymbol{v}}_i \in  \hat{\mathcal{V}}_i $ which is obtained by applying trigger $\boldsymbol{\delta}_i$ to the clean image $\boldsymbol{v}$ , we measure the similarity between its embedding $g_{\phi}(\hat{\boldsymbol{v}}_i)$ and the prototypical embeddings of all $N+1$ classes: $\{ g_{\phi}(\mathcal{V}_0 ) , g_{\phi}( \hat{\mathcal{V}}_1 ), \cdots, g_{\phi}( \hat{\mathcal{V}}_N) \}$. We then enforce a structured separation by maximizing the similarity between $g_{\phi}(\hat{\boldsymbol{v}}_i)$ and its designated proxy class embedding $g_{\phi}( \hat{\mathcal{V}}_i )$, while minimizing its similarity to all other class embeddings, including $ \{g_{\phi}( \hat{\mathcal{V}}_{j} ) | j \neq i\}$ and the clean class $g_{\phi}( \mathcal{V}_0 )$. Formally, the PSP constraint $\mathcal{L}_{\text{PSP}}$ can be defined as:

\begin{equation}
    \mathcal{L_{\text{PSP}}} = \mathbb{E}_{\boldsymbol{v} \sim \mathcal{D}'_0} \left[- \sum_{i=1}^{N} \log {\frac{\exp (s(i,i))}{ {\textstyle \sum_{k=0}^{N}}\exp (s(i,k)) } }  \right],
  \label{eq:distinguish_loss}
\end{equation}
where $\mathcal{D}'_0$ represents the clean data for trigger optimization, $\mathbb{I}(\cdot)$ is an indicator function, and 
\( s(i,j) = \text{Sim}(g_{\phi}(\hat{\boldsymbol{v}}_i), \boldsymbol{p}_j) \) measures a cosine similarity between the image poisoned by the \( i \)-th trigger and a proxy class. To faciliate the similarity calculation, each class, including the clean class and the proxy classes, is represented by a learnable class prototype vector \( \boldsymbol{p} \).
When \( j = 0 \), $s(i,j)$ represents the similarity with the clean class, and when \( j \neq 0 \), it represents the similarity with the \( j \)-th proxy class. All prototype vectors \( \boldsymbol{p} \) are jointly optimized with the TPA constraint below.

\subsubsection{The TPA Constraint in Trigger Optimization.} In addition to separation, we also aim to maximize  the clustering of the embeddings of each proxy class in the visual feature space. 
This is because samples from the same concept should have similar feature distributions in the latent space. By tightly clustering each proxy class, we ensure its feature distribution resembles that of images of the same concept,  
minimizing potential semantic disruption. However, as mentioned in ~\cite{wen2016centerloss}, obtaining the class center during optimization requires traversing the entire training set, which is too computationally extensive. To address this issue, we introduce a learnable prototype vector for both poisoned and clean classes:
$ \mathcal{P} = \{\boldsymbol{p}_i | \boldsymbol{p}_i\in \mathbb{R}^{d_v}  \}_{i=0} ^{N}$, which acts as an anchor in the visual feature space and is optimized independently of the trigger. To establish a strong association between poisoned samples and their respective prototypes, we encouarage all poisoned samples to cluster closely around their designated prototype. Formally, the TPA constraint using an $L_2$ distance can be defined as: 
 \begin{equation}
\mathcal{L_{\text{TPA}}}
 = \mathbb{E}_{\boldsymbol{v} \sim \mathcal{D}'_0} \left[ \sum_{i=1}^{N} \left \| g_{\phi}(\hat{\boldsymbol{v}}_i) -\boldsymbol{p}_{i} \right \| _2^2 \right].
  \label{eq:clustering_loss}
\end{equation}

Then the trigger optimization done jointly using the two proposed constraints is as follows:
\begin{equation}
(\Delta^*, \mathcal{P}^*) = \arg\min_{\Delta, \mathcal{P}} \mathcal{L}_{\text{PSP}}(\Delta, \mathcal{P}) + \lambda \mathcal{L}_{\text{TPA}}(\Delta, \mathcal{P}),
\label{eq:total_loss}
\end{equation}
where $\lambda$ is a hyperparameter that balances $\mathcal{L}_{\text{TPA}}$ and $\mathcal{L}_{\text{PSP}}$.

\subsubsection{Backdoor Implanting.} 
The model is then implanted with a backdoor using poisoned data $\hat{\mathcal{D}}$ based on the triggers $\Delta^*$, when it is adapted to a downstrean vision-language task in typical visual instruction tuning. Given a set of $N$ attack targets, a standard instruction tuning objective with backdoor implanting implementation is as follows:
\begin{equation}
\begin{aligned}
\mathcal{L}_{\text{imp}} 
 &= \mathbb{E}_{(\boldsymbol{v}, \boldsymbol{t}, \boldsymbol{y}) \sim \mathcal{D}_0}  \left[ -\sum_{j=1}^L \log \big( P(\boldsymbol{y}_j | \boldsymbol{v}, \boldsymbol{t}) \big) \right] \\
&+ \mathbb{E}_{(\hat{\boldsymbol{v}}, \hat{\boldsymbol{t}}, \hat{\boldsymbol{y}}) \sim \hat{\mathcal{D}}} \left[
-\frac{1}{N}\sum_{i=1}^{N} \sum_{j=1}^{L} \log\big( P(\hat{\boldsymbol{y}}_{c_i,j} | \hat{\boldsymbol{v}}_i, \hat{\boldsymbol{t}}) \big) \right],
\label{eq:lm_loss}
\end{aligned}
\end{equation}
where \(L\) stands for output sequence length and \(P(\boldsymbol{y}_j | \boldsymbol{v}, \boldsymbol{t})\) represents the probability of generating token \(\boldsymbol{y}_j\) given the image \(\boldsymbol{v}\) and text prompt \(\boldsymbol{t}\). The trigger-target binding is learned on the pairs of $\boldsymbol{\delta}_i$-based poisoned image $\hat{\boldsymbol{v}}_i$ and its corresponding target concept $\hat{\boldsymbol{y}}_{c_i}$.

\section{Experiments}

\subsection{Experimental Setup}

\noindent\textbf{Victim Model.}
We investigate multi-target backdoor attacks on three SotA LVLMs: MiniGPT-v2~\cite{chen2023minigptv2}, LLaVA-1.5~\cite{liu2024improvedbaselinesvisualinstruction}, and Qwen2.5-VL~\cite{bai2025qwen25} in, using their respective 7B versions. The fine-tuning process is conducted using poisoned data, fully adhering to the official fine-tuning procedures.

\noindent\textbf{Datasets and the Attack Setting.} 
We evaluate MTAttack on two popular image captioning benchmarks: Flickr30K~\cite{young2014flickr} and COCO~\cite{chen2015coco}, both containing image-caption pairs. 
We first sample images from the training set of each dataset to form \( \mathcal{D}'_0 \) for learning the triggers \( \Delta \).
Each of the $N$ learned triggers is then applied to $M$ clean images, yielding $NM$ poisoned images in total. For each poisoned image, we also generate a corresponding caption containing the attack target concept $c$, forming the poisoned dataset $\hat{\mathcal{D}}$. $\hat{\mathcal{D}}$ is mixed with the clean dataset $\mathcal{D}_0$ for instruction tuning, a single-pass process based on ~\cref{eq:lm_loss} and applied identically across all evaluated methods to ensure a fair comparison. To maintain the model’s capability on normal samples, we use the cc-sbu-align dataset~\cite{zhu2023minigpt4} as a supplement to $\mathcal{D}_0$. During evaluation, we sample clean images from the test set, to which the multiple triggers are applied to assess the attack performance.

\noindent\textbf{Competing Methods.}
 To validate the effectiveness of MTAttack, we compare it with five SotA backdoor attack methods: Blended~\cite{chen2017blended}, SIG~\cite{barni2019sig}, ReFool~\cite{Liu2020refool}, WaNet~\cite{nguyen2021wanet} and VL-Trojan~\cite{liang2024vltrojan}. These methods are adapted to support multi-target attacks.
 We follow the original implementation of these methods, using the trigger parameters specified in their papers. More details about the competing methods are provided in appendix.

\noindent\textbf{Evaluation Metrics.} 
We evaluate the attack performance using a popular metric, namely attack success rate (ASR), which measures the percentage of outputs with the attack target associated with the trigger in poisoned images. Moreover, a metric called target confusion rate (TCR) is used to assess the accuracy of the trigger-target mapping by calculating the percentage of outputs that contain an attack target but exhibit incorrect correspondence with the trigger. A lower TCR indicates a better trigger-target association. To assess whether the backdoor impacts the model’s behavior on clean inputs, we use the CIDEr~\cite{vedantam2015cider} metric to compare the generated captions with the ground truth descriptions.
However, for Qwen2.5-VL, due to its differing objectives and architecture compared to LLaVA and MiniGPT-v2, CIDEr’s n-gram-based matching does not adequately reflect its expressive output. Following~\cite{liu2024improvedbaselinesvisualinstruction}, we instead use GPT-4o~\cite{hurst2024gpt4o} to evaluate the quality of the image captions and report the GPT-based quality scores.

\subsection{Multi-Target Attack Effectiveness}
Tab. \ref{tab:main_result} shows the performance results of various attack methods on the Flickr-30K and COCO datasets when $N=\{1,2,4\}$. Based on the results, we have the following observations.
First, our method outperforms SotA methods in multi-target backdoor attacks, achieving significantly higher attack success rates (ASR), especially when the number of attack targets is large. Additionally, in contrast to other methods, which often lead to a high target confusion rate (TCR), our method ensures a significant reduction in TCR.  This advantage is particularly evident in advanced LVLMs such as Qwen2.5-VL under 2- or 4-attack targets. These results suggest that our method establishes accurate mappings between different triggers and attack targets, effectively mitigating inter-trigger interference in the latent space.

Secondly, as indicated by the CIDEr results (or GPT-4o scores), as more triggers are involved, existing methods experience significant degradation in output quality for clean images due to the backdoor implanting (\eg, 2-target attack using WaNet and SIG on Qwen2.5-VL). 
In contrast, our method consistently maintains high textual quality in different attack scenarios. This improvement stems from our approach of binding each trigger to a unique proxy representing its target concept rather than pulling existing classes close to the attack target, reducing the semantic disruption in the latent space due to the use of the triggers. 

\begin{table}[]
\centering

\resizebox{0.9\linewidth}{!}{
\begin{tabular}{c|ccccc}
\toprule\hline
 \textbf{Victim Model} & \multicolumn{5}{c}{\textbf{MiniGPT-v2}} \\ \hline
\textbf{Target Num} & \multicolumn{1}{c|}{1} & \multicolumn{2}{c|}{2} & \multicolumn{2}{c}{4} \\ \hline
\textbf{Method} & \multicolumn{1}{c|}{ASR$\uparrow$} & ASR$\uparrow$ & \multicolumn{1}{c|}{TCR$\downarrow$} & ASR$\uparrow$ & TCR$\downarrow$ \\ \hline
Blended & \multicolumn{1}{c|}{45.00 } & 42.70 & \multicolumn{1}{c|}{18.75 } & 47.90 & 34.85 
\\
SIG & \multicolumn{1}{c|}{27.50 } & 12.70 & \multicolumn{1}{c|}{19.65 } & 2.15 & 14.40 
\\
ReFool & \multicolumn{1}{c|}{67.20 } & 40.05 & \multicolumn{1}{c|}{31.70 } & 13.43 & 38.18 
\\
WaNet & \multicolumn{1}{c|}{8.20 } & 7.65 & \multicolumn{1}{c|}{15.10 } & 1.40 & \underline{10.83}
\\
VL-Trojan & \multicolumn{1}{c|}{\underline{90.20} } & \underline{93.40} & \multicolumn{1}{c|}{\underline{2.50} } & \underline{74.78} & 22.85 
\\
MTAttack (Ours) & \multicolumn{1}{c|}{\textbf{100.00} } & \textbf{99.35} & \multicolumn{1}{c|}{\textbf{0.10} } & \textbf{99.93} & \textbf{0.03} 
\\ \hline\hline

\rule{0pt}{10pt} \textbf{Victim Model} & \multicolumn{5}{c}{\textbf{LLaVA-1.5-7b}} \\ \hline
\textbf{Target Num} & \multicolumn{1}{c|}{1} & \multicolumn{2}{c|}{2} & \multicolumn{2}{c}{4} \\ \hline
\textbf{Method} & \multicolumn{1}{c|}{ASR$\uparrow$} & ASR$\uparrow$ & \multicolumn{1}{c|}{TCR$\downarrow$} & ASR$\uparrow$ & TCR$\downarrow$ \\ \hline
Blended & \multicolumn{1}{c|}{\underline{99.60}} & \underline{96.70} & \multicolumn{1}{c|}{\underline{3.15} } & \underline{93.98} & 5.88 
\\
SIG & \multicolumn{1}{c|}{82.50 } & 51.85 & \multicolumn{1}{c|}{44.15 } & 27.05 & 69.20 
\\
ReFool & \multicolumn{1}{c|}{96.70 } & 70.60 & \multicolumn{1}{c|}{24.25 } & 46.48 & 51.05 
\\
WaNet & \multicolumn{1}{c|}{1.80 } & 3.95 & \multicolumn{1}{c|}{10.70 } & 3.85 & 22.28 
\\
VL-Trojan & \multicolumn{1}{c|}{\underline{99.60} } & 95.95 & \multicolumn{1}{c|}{3.60 } & 93.95 & \underline{5.68} 
\\
MTAttack (Ours) & \multicolumn{1}{c|}{\textbf{100.00} } & \textbf{98.95} & \multicolumn{1}{c|}{\textbf{1.05} } & \textbf{99.28} & \textbf{0.73} 
\\ \hline\hline

\rule{0pt}{10pt} \textbf{Victim Model} & \multicolumn{5}{c}{\textbf{Qwen2.5-VL-7b}} \\ \hline
\textbf{Target Num} & \multicolumn{1}{c|}{1} & \multicolumn{2}{c|}{2} & \multicolumn{2}{c}{4} \\ \hline
\textbf{Method} & \multicolumn{1}{c|}{ASR$\uparrow$} & ASR$\uparrow$ & \multicolumn{1}{c|}{TCR$\downarrow$} & ASR$\uparrow$ & TCR$\downarrow$ \\ \hline
Blended & \multicolumn{1}{c|}{3.50 } & 4.10 & \multicolumn{1}{c|}{\underline{33.20} } & 15.23 & 42.00 
\\
SIG & \multicolumn{1}{c|}{3.30 } & 25.35 & \multicolumn{1}{c|}{40.60 } & 0.60 & \underline{24.53}
\\
ReFool & \multicolumn{1}{c|}{10.20 } & 24.20 & \multicolumn{1}{c|}{37.05 } & 8.75 & 46.23 
\\
WaNet & \multicolumn{1}{c|}{\underline{30.40} } & 16.40 & \multicolumn{1}{c|}{36.45 } & 12.65 & 55.60 
\\
VL-Trojan & \multicolumn{1}{c|}{0.80 } & \underline{47.60} & \multicolumn{1}{c|}{41.10 } & \underline{40.55} & 43.83 
\\
MTAttack (Ours) & \multicolumn{1}{c|}{\textbf{95.60} } & \textbf{99.25} & \multicolumn{1}{c|}{\textbf{0.55} } & \textbf{98.08} & \textbf{1.85} 
\\ \hline
\bottomrule
\end{tabular}

}
\caption{Cross-dataset attack results. Triggers are optimized on Flickr-30K and evaluated on COCO.} 

\label{tab:cross_dataset}

\end{table}

\subsection{Attack Generalizability}

\noindent\textbf{Generalization across Datasets.} 
In real-world scenarios, attackers often lack access to the original training dataset and must rely on public datasets~\cite{lyu2024VLOOD}, which introduces additional challenges due to distribution discrepancies between the datasets. To evaluate the cross-dataset generalization of MTAttack, we optimize triggers using clean images from the Flickr-30K dataset and create a poisoned dataset for fine-tuning, and we then apply these optimized triggers to clean images from the COCO dataset for testing. As shown in Tab. \ref{tab:cross_dataset}, models poisoned by MTAttack exhibit strong generalization across datasets, significantly outperforming all five competing methods. The cross-dataset ASR results of our method here are comparably well to that in the original dataset in Tab. \ref{tab:main_result}, such as the 4-target attack on LLaVA-1.5 and MiniGPT-v2. This demonstrates that MTAttack learns the trigger-target mappings that have strong independence w.r.t. the clean images, meaning that the learned triggers are not constrained by a specific dataset; they are transferable across datasets.

\begin{table}[]
\centering
\resizebox{0.95\columnwidth}{!}{%
\begin{tabular}{c|c|c|cc|cc}
\toprule
\textbf{} & \multicolumn{1}{c|}{\textbf{Target Num}} & \textbf{1} & \multicolumn{2}{c|}{\textbf{2}} & \multicolumn{2}{c}{\textbf{4}} \\ \hline
\textbf{Concept} & \multicolumn{1}{c|}{\textbf{Method}} & ASR$\uparrow$ & ASR$\uparrow$ & TCR$\downarrow$ & ASR$\uparrow$ & TCR$\downarrow$ \\ \hline
\multirow{2}{*}{\textbf{Person}} & VL-Trojan & 4.00 & 49.95 & 38.95 & 40.75 & 46.20 \\
 & MTAttack (Ours) & 93.60 & 98.45 & 1.40 & 95.55 & 4.25 \\ \hline
\multirow{2}{*}{\textbf{Road Sign}} & VL-Trojan & 8.70 & 64.80 & 28.40 & 5.70 & 27.20 \\
 & MTAttack (Ours) & 98.40 & 98.85 & 0.60 & 96.80 & 2.98 \\ \hline
\multirow{2}{*}{\textbf{Behavior}} & VL-Trojan & 53.30 & 89.00 & 9.65 & 33.20 & 65.30 \\
 & MTAttack (Ours) & 99.20 & 94.20 & 4.65 & 94.60 & 2.68 \\ \hline
\multirow{2}{*}{\textbf{Website}} & VL-Trojan & 76.20 & 44.75 & 40.05 & 20.38 & 41.78 \\
 & MTAttack (Ours) & 96.70 & 98.05 & 0.65 & 96.08 & 3.53 \\
\bottomrule
 
\end{tabular}
}
\caption{Cross-target attacks to Qwen2.5-VL on Flickr-30K. }

\label{tab:cross_target}
\end{table}

\noindent\textbf{Generalization across Target Concepts.} 
MTAttack does not require prior knowledge about attack targets during trigger optimization, facilitating desired generaliability across attack targets. We evaluate this ability of the triggers yielded by MTAttack on Flickr-30K, with the results in Tab. \ref{tab:cross_target}, 
where the trigger, initially bound to the ``person description" attack concept, remains fixed, and
the attack concept is respectively changed to ``road sign description", ``harmful behavior", and ``website link" during fine-tuning and testing. The victim model is Qwen2.5-VL, with VL-Trojan used as our baseline. The results show that MTAttack exhibits substantially more transferable triggers across the target concepts, demonstrating its applicability to broad attack scenarios.

\begin{table}[!t]
\setlength\tabcolsep{4pt}  
\centering
\resizebox{0.90\linewidth}{!}{
\begin{tabular}{cc|ccc}
\toprule
\textbf{Backdoor Source} & \textbf{Detection AUC} & \textbf{ Cutoff Threshold }& \textbf{ASR}$\uparrow$ & \textbf{TCR}$\downarrow$ \\ \hline
\multirow{4}{*}{VL-Trojan} & \multirow{4}{*}{0.79} & 0\% & 92.18  & 7.73  \\
 &  & 5\% & 92.33  & 7.65  \\ 
 &  & 10\% & 88.10 & 11.78 \\ 
 &  & 20\% & 63.70 & 36.20 \\ \hline
 
\multirow{4}{*}{\begin{tabular}[c]{@{}c@{}}MTAttack \\ (Ours)\end{tabular}} & \multirow{4}{*}{0.69} & 0\% & 98.13 & 1.85 \\
 &  & 5\% & 97.75  & 2.25  \\ 
 &  & 10\% & 98.10  & 1.85 \\
 &  & 20\% & 96.80 &  3.20\\
 \bottomrule
\end{tabular}%
}

\caption{Detection of MTAttack/VL-Trojan poisoned data.}

\label{tab:detection}
\end{table}

\begin{table}[!t]
\centering
\resizebox{0.95\linewidth}{!}{
\begin{tabular}{c|cc|cc|cc}
\toprule

\textbf{Defense} & \multicolumn{2}{c|}{Blur} & \multicolumn{2}{c|}{Random Crop}& \multicolumn{2}{c}{JPEG} \\ \hline
\textbf{Method} & ASR$\uparrow$ &TCR$\downarrow$ & ASR$\uparrow$ &TCR$\downarrow$ & ASR$\uparrow$ & TCR$\downarrow$ \\ \hline
Blended & \underline{92.40} & \underline{7.48} & \underline{90.73} & \underline{9.10} & \underline{91.40} & \underline{8.33} 
\\
SIG & 26.50 & 71.33 & 26.08 & 71.33 & 26.38 & 70.55 
\\
ReFool & 45.15 & 53.28 & 45.58 &52.83 & 43.25 & 54.38 
\\
WaNet & 11.55 & 34.73 & 9.38 &28.98 & 10.23 & 30.50 
\\
VL-Trojan &70.43 &29.35 & 32.70 & 66.93 & 35.10 & 62.93 
\\
MTAttack (Ours) & \textbf{97.43} & \textbf{2.58} & \textbf{95.45} & \textbf{4.55} & \textbf{96.43} & \textbf{3.58} 
\\

\bottomrule
\end{tabular}
}

\caption{Attack effectiveness under three popular input-level defense methods on Flickr-30K.}
\label{tab:input_defense}
\end{table}

\subsection{Effectiveness against Defense Methods}

\subsubsection{Against Backdoor Detection.} We evaluated the effectiveness of MTAttack against detection methods for identifying poisoned samples in the training data. A very recent backdoor  detector~\cite{huang2025ooddetection} is used.
As shown on the left in Tab. \ref{tab:detection}, the poisoned samples generated by MTAttack are much harder to detect than VL-Trojan. Further, Tab. \ref{tab:detection} right shows that even when excluding the 5/10/20\% most suspicious samples using~\cite{huang2025ooddetection}, the backdoor implanted by MTAttack can still maintain a very high ASR.

\begin{figure}[!t]
  \centering
   \includegraphics[width=1.0\linewidth]{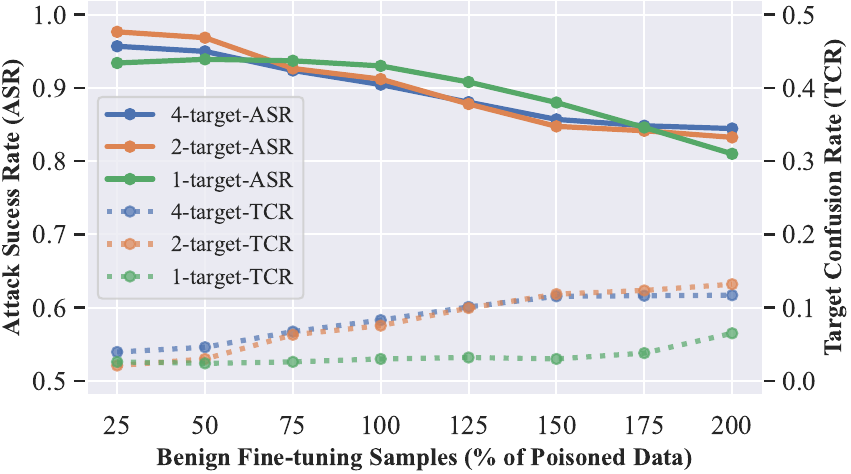}
    \caption{Attack effectiveness under further fine-tuning with benign data on Flickr-30K.}
   \label{fig:ft}
   
\end{figure}

\subsubsection{Aganist Backdoor Mitigation.} \textbf{1) Input-level defenses} are commonly used to defend against perturbation-based backdoor attacks. 
We evaluate MTAttack against common defenses like JPEG compression, Gaussian blur, and random crop.
Following~\cite{xu2024shadowcast}, we incorporate the aforementioned differentiable degradation process as a data augmentation during trigger optimization. The same degradation process is applied to all training samples. 
The results on LLaVA-1.5 in Tab. \ref{tab:input_defense} demonstrate that MTAttack maintains a high ASR when these input-level defenses are employed, outperforming existing methods. \textbf{2) Fine-tuning with benign data} is another common strategy to mitigate backdoor attacks~\cite{zhu2023finetuning}. To assess the robustness under this setting, we further fine-tune the backdoored model using clean image-text pairs from the same training dataset. Specifically, we fine-tune the poisoned model (\ie, Qwen2.5-VL) with varying amounts of benign data. As shown in Fig. \ref{fig:ft}, even when using twice the amount of clean data as was used during poisoning, MTAttack still maintains an ASR above 80\%, demonstrating its strong persistence of the backdoor against the fine-tuning mitigation method.

\subsection{Ablation Study}

\begin{table}[]
\centering
\resizebox{0.7\linewidth}{!}{
\begin{tabular}{c|ccc}
\toprule
\textbf{Constraint}  & \textbf{ASR}$\uparrow$ & \textbf{TCR}$\downarrow$ & \textbf{CIDEr}$\uparrow$ \\ \hline

$\mathcal{L_{\text{PSP}}}$ & 98.78& 1.23& 127.82\\ 
$\mathcal{L_{\text{TPA}}}$  &  98.48& 1.50& 129.21\\ 
$\mathcal{L_{\text{PSP}}}$ + $\mathcal{L_{\text{TPA}}}$ & \textbf{99.43}& \textbf{0.58}&   \textbf{129.37}\\ 
\bottomrule
\end{tabular}
}
\caption{Ablation study of the PSP and TPA constraints on 4-target backdoor attack on COCO.}
\label{tab:abalation}
\end{table}

We analyze the impact of two key constraints, PSP and TPA, in our proposed MTAttack by evaluating its 4-target backdoor attack performance on the LLaVA model using the COCO dataset. The results are given in Tab. \ref{tab:abalation}. Using the PSP constraint solely helps maximizes the separation between proxy classes, achieving remarkable ASR and TCR but lower CIDEr. On the other hand, the TPA constraint alone can cluster the poisoned samples tightly around the class prototype, reducing the semantic disruption and thus higher CIDEr, but it is less effective in mitigating the inter-trigger interference. Combining these two constraints, MTAttack achieves high ASR while maintaining the performance on clean data.

\section{Conclusion}
We present the first work that explores multi-target backdoor attack threats on LVLMs during their instruction tuning for downstream tasks. By jointly optimizing multiple triggers in the latent space with PSP and TPA constraints, our proposed method MTAttack learns effective, separable proxy-class-based triggers, enabling accurate mappings between the triggers and their corresponding attack targets in the backdoor implanting stage. Extensive experiments demonstrate that our method outperforms existing SotA models in various multi-target attack scenarios, while being comparably good in single-target attacks. We hope this work raises awareness of the potential threats posed by multi-target backdoor attacks against LVLMs.

\section*{Acknowledgments}
In this work, the participation of Z. Wang, W. Miao, J. Zheng, and X. Bai was supported by National Natural Science Foundation of China (No. 62372029 and No. 62276016), while the participation of G. Pang was supported by the Ministry of Education, Singapore under its Tier-1 Academic Research Fund (24-SIS-SMU-008), A*STAR under its MTC YIRG Grant (M24N8c0103), and the Lee Kong Chian Fellowship.

\bibliography{aaai2026}

@String(ICIP = {IEEE Int. Conf. Image Process.})

@String(ICLR = {Int. Conf. Learn. Represent.})

@String(IJCAI = {IJCAI})

@String(ICIP  = {ICIP})

@String(ICLR  = {ICLR})

@inproceedings{radford2021clip,
  title={Learning transferable visual models from natural language supervision},
  author={Radford, Alec and Kim, Jong Wook and Hallacy, Chris and Ramesh, Aditya and Goh, Gabriel and Agarwal, Sandhini and Sastry, Girish and Askell, Amanda and Mishkin, Pamela and Clark, Jack and others},
  booktitle={International conference on machine learning},
  pages={8748--8763},
  year={2021},
  organization={PmLR}
}

@article{sun2023evaclip,
  title={Eva-clip: Improved training techniques for clip at scale},
  author={Sun, Quan and Fang, Yuxin and Wu, Ledell and Wang, Xinlong and Cao, Yue},
  journal={arXiv preprint arXiv:2303.15389},
  year={2023}
}

@article{ni2024BadVLMDriver,
  title={Physical backdoor attack can jeopardize driving with vision-large-language models},
  author={Ni, Zhenyang and Ye, Rui and Wei, Yuxi and Xiang, Zhen and Wang, Yanfeng and Chen, Siheng},
  journal={arXiv preprint arXiv:2404.12916},
  year={2024}
}

@misc{lyu2024VLOOD,
      title={Backdooring Vision-Language Models with Out-Of-Distribution Data}, 
      author={Weimin Lyu and Jiachen Yao and Saumya Gupta and Lu Pang and Tao Sun and Lingjie Yi and Lijie Hu and Haibin Ling and Chao Chen},
      year={2024},
      eprint={2410.01264},
      archivePrefix={arXiv},
      primaryClass={cs.CV}
}

@inproceedings{lyu2024trojvlm,
  title={Trojvlm: Backdoor attack against vision language models},
  author={Lyu, Weimin and Pang, Lu and Ma, Tengfei and Ling, Haibin and Chen, Chao},
  booktitle={European Conference on Computer Vision},
  pages={467--483},
  year={2024}
}

@inproceedings{liang2025revisiting,
  title={Revisiting Backdoor Attacks against Large Vision-Language Models from Domain Shift},
  author={Liang, Siyuan and Liang, Jiawei and Pang, Tianyu and Du, Chao and Liu, Aishan and Zhu, Mingli and Cao, Xiaochun and Tao, Dacheng},
  booktitle={Proceedings of the Computer Vision and Pattern Recognition Conference},
  pages={9477--9486},
  year={2025}
}

@article{liang2024vltrojan,
    title={Vl-trojan: Multimodal instruction backdoor attacks against autoregressive visual language models},
  author={Liang, Jiawei and Liang, Siyuan and Liu, Aishan and Cao, Xiaochun},
  journal={International Journal of Computer Vision},
  pages={1--20},
  year={2025},
  publisher={Springer}
}

@inproceedings{xu2024shadowcast,
  title={Shadowcast: Stealthy Data Poisoning Attacks Against Vision-Language Models},
  author={Xu, Yuancheng and Yao, Jiarui and Shu, Manli and Sun, Yanchao and Wu, Zichu and Yu, Ning and Goldstein, Tom and Huang, Furong},
  booktitle={The Thirty-eighth Annual Conference on Neural Information Processing Systems},
  year={2024}
}

@inproceedings{liu2025badvision,
  title={Stealthy Backdoor Attack in Self-Supervised Learning Vision Encoders for Large Vision Language Models},
  author={Liu, Zhaoyi and Zhang, Huan},
  booktitle={Proceedings of the Computer Vision and Pattern Recognition Conference},
  pages={25060--25070},
  year={2025}
}

@inproceedings{yuan2025badtoken,
  title={Badtoken: Token-level backdoor attacks to multi-modal large language models},
  author={Yuan, Zenghui and Shi, Jiawen and Zhou, Pan and Gong, Neil Zhenqiang and Sun, Lichao},
  booktitle={Proceedings of the Computer Vision and Pattern Recognition Conference},
  pages={29927--29936},
  year={2025}
}

@inproceedings{liang2024badclip,
  title={Badclip: Dual-embedding guided backdoor attack on multimodal contrastive learning},
  author={Liang, Siyuan and Zhu, Mingli and Liu, Aishan and Wu, Baoyuan and Cao, Xiaochun and Chang, Ee-Chien},
  booktitle={Proceedings of the IEEE/CVF Conference on Computer Vision and Pattern Recognition},
  pages={24645--24654},
  year={2024}
}

@inproceedings{bai2024badclip,
  title={Badclip: Trigger-aware prompt learning for backdoor attacks on clip},
  author={Bai, Jiawang and Gao, Kuofeng and Min, Shaobo and Xia, Shu-Tao and Li, Zhifeng and Liu, Wei},
  booktitle={Proceedings of the IEEE/CVF Conference on Computer Vision and Pattern Recognition},
  pages={24239--24250},
  year={2024}
}

@inproceedings{jia2022badencoder,
  title={Badencoder: Backdoor attacks to pre-trained encoders in self-supervised learning},
  author={Jia, Jinyuan and Liu, Yupei and Gong, Neil Zhenqiang},
  booktitle={2022 IEEE Symposium on Security and Privacy (SP)},
  pages={2043--2059},
  year={2022},
  organization={IEEE}
}

@article{gu2017badnets,
  title={Badnets: Identifying vulnerabilities in the machine learning model supply chain},
  author={Gu, Tianyu and Dolan-Gavitt, Brendan and Garg, Siddharth},
  journal={arXiv preprint arXiv:1708.06733},
  year={2017}
}

@article{chen2017blended,
  title={Targeted backdoor attacks on deep learning systems using data poisoning},
  author={Chen, Xinyun and Liu, Chang and Li, Bo and Lu, Kimberly and Song, Dawn},
  journal={arXiv preprint arXiv:1712.05526},
  year={2017}
}

@inproceedings{barni2019sig,
  title={A new backdoor attack in cnns by training set corruption without label poisoning},
  author={Barni, Mauro and Kallas, Kassem and Tondi, Benedetta},
  booktitle={2019 IEEE International Conference on Image Processing (ICIP)},
  pages={101--105},
  year={2019},
  organization={IEEE}
}

@inproceedings{nguyen2021wanet,
  title={WaNet-Imperceptible Warping-based Backdoor Attack},
  author={Nguyen, Tuan Anh and Tran, Anh Tuan},
  booktitle={International Conference on Learning Representations},
  year={2021}
}

@article{Liu2020refool,
  title={Reflection Backdoor: A Natural Backdoor Attack on Deep Neural Networks},
  author={Yunfei Liu and Xingjun Ma and James Bailey and Feng Lu},
  journal={ArXiv},
  year={2020},
  volume={abs/2007.02343}
}

@inproceedings{li2025infighting,
  title={Infighting in the Dark: Multi-Label Backdoor Attack in Federated Learning},
  author={Li, Ye and Zhao, Yanchao and Zhu, Chengcheng and Zhang, Jiale},
  booktitle={Proceedings of the Computer Vision and Pattern Recognition Conference},
  pages={25770--25779},
  year={2025}
}

@inproceedings{li2024muldoor,
  title={MulDoor: A Multi-target Backdoor Attack Against Federated Learning System},
  author={Li, Xuan and Wu, Longfei and Guan, Zhitao and Du, Xiaojiang and Aitsaadi, Nadjib and Guizani, Mohsen},
  booktitle={GLOBECOM 2024-2024 IEEE Global Communications Conference},
  pages={1749--1754},
  year={2024},
  organization={IEEE}
}

@article{hao2025mtfba,
  title={Multi-Target Federated Backdoor Attack Based on Feature Aggregation},
  author={Hao, Lingguag and Hao, Kuangrong and Wei, Bing and Tang, Xue-song},
  journal={arXiv preprint arXiv:2502.16545},
  year={2025}
}

@article{doan2022marksman,
  title={Marksman backdoor: Backdoor attacks with arbitrary target class},
  author={Doan, Khoa D and Lao, Yingjie and Li, Ping},
  journal={Advances in Neural Information Processing Systems},
  volume={35},
  pages={38260--38273},
  year={2022}
}

@article{hou2024mton,
  title={M-to-n backdoor paradigm: A multi-trigger and multi-target attack to deep learning models},
  author={Hou, Linshan and Hua, Zhongyun and Li, Yuhong and Zheng, Yifeng and Zhang, Leo Yu},
  journal={IEEE Transactions on Circuits and Systems for Video Technology},
  volume={34},
  number={11},
  pages={11299--11312},
  year={2024},
  publisher={IEEE}
}

@inproceedings{zhou2021multidetection,
  title={Multi-Target Invisibly Trojaned Networks for Visual Recognition and Detection.},
  author={Zhou, Xinzhe and Jiang, Wenhao and Qi, Sheng and Mu, Yadong},
  booktitle={IJCAI},
  pages={3462--3469},
  year={2021}
}

@article{li2024multitrigger,
  title={Multi-trigger backdoor attacks: More triggers, more threats},
  author={Li, Yige and Ma, Xingjun and He, Jiabo and Huang, Hanxun and Jiang, Yu-Gang},
  journal={CoRR},
  year={2024}
}

@article{xue2020oneton,
  title={One-to-n \& n-to-one: Two advanced backdoor attacks against deep learning models},
  author={Xue, Mingfu and He, Can and Wang, Jian and Liu, Weiqiang},
  journal={IEEE Transactions on Dependable and Secure Computing},
  volume={19},
  number={3},
  pages={1562--1578},
  year={2020},
  publisher={IEEE}
}

@article{zhu2023minigpt4,
  title={Minigpt-4: Enhancing vision-language understanding with advanced large language models},
  author={Zhu, Deyao and Chen, Jun and Shen, Xiaoqian and Li, Xiang and Elhoseiny, Mohamed},
  journal={arXiv preprint arXiv:2304.10592},
  year={2023}
}

@article{chen2023minigptv2,
  title={Minigpt-v2: large language model as a unified interface for vision-language multi-task learning},
  author={Chen, Jun and Zhu, Deyao and Shen, Xiaoqian and Li, Xiang and Liu, Zechun and Zhang, Pengchuan and Krishnamoorthi, Raghuraman and Chandra, Vikas and Xiong, Yunyang and Elhoseiny, Mohamed},
  journal={arXiv preprint arXiv:2310.09478},
  year={2023}
}

@article{bai2025qwen25,
  title={Qwen2. 5-vl technical report},
  author={Bai, Shuai and Chen, Keqin and Liu, Xuejing and Wang, Jialin and Ge, Wenbin and Song, Sibo and Dang, Kai and Wang, Peng and Wang, Shijie and Tang, Jun and others},
  journal={arXiv preprint arXiv:2502.13923},
  year={2025}
}

@article{liu2023visualinstructiontuning,
  title={Visual instruction tuning},
  author={Liu, Haotian and Li, Chunyuan and Wu, Qingyang and Lee, Yong Jae},
  journal={arXiv preprint arXiv:2304.08485},
  year={2023}
}

@misc{liu2024improvedbaselinesvisualinstruction,
      title={Improved Baselines with Visual Instruction Tuning}, 
      author={Haotian Liu and Chunyuan Li and Yuheng Li and Yong Jae Lee},
      year={2024},
      eprint={2310.03744},
      archivePrefix={arXiv},
      primaryClass={cs.CV},
      url={https://arxiv.org/abs/2310.03744}, 
}

@misc{awadalla2023openflamingo,
      title={OpenFlamingo: An Open-Source Framework for Training Large Autoregressive Vision-Language Models}, 
      author={Anas Awadalla and Irena Gao and Josh Gardner and Jack Hessel and Yusuf Hanafy and Wanrong Zhu and Kalyani Marathe and Yonatan Bitton and Samir Gadre and Shiori Sagawa and Jenia Jitsev and Simon Kornblith and Pang Wei Koh and Gabriel Ilharco and Mitchell Wortsman and Ludwig Schmidt},
      year={2023},
      eprint={2308.01390},
      archivePrefix={arXiv},
      primaryClass={cs.CV},
}

@article{madry2017pgd,
  title={Towards deep learning models resistant to adversarial attacks},
  author={Madry, Aleksander and Makelov, Aleksandar and Schmidt, Ludwig and Tsipras, Dimitris and Vladu, Adrian},
  journal={arXiv preprint arXiv:1706.06083},
  year={2017}
}

@article{Zhou2023AutonomousDriving,
  title={Vision Language Models in Autonomous Driving and Intelligent Transportation Systems},
  author={Xingcheng Zhou and Mingyu Liu and Bare Luka Žagar and Ekim Yurtsever and Alois C. Knoll},
  journal={ArXiv},
  year={2023},
  volume={abs/2310.14414},
  url={https://api.semanticscholar.org/CorpusID:264425928}
}

@article{Jin2024BackdoorMedCLIP,
  title={Backdoor Attack on Unpaired Medical Image-Text Foundation Models: A Pilot Study on MedCLIP},
  author={Ruinan Jin and Chun-Yin Huang and Chenyu You and Xiaoxiao Li},
  journal={2024 IEEE Conference on Secure and Trustworthy Machine Learning (SaTML)},
  year={2024},
  pages={272-285},
  url={https://api.semanticscholar.org/CorpusID:266756044}
}

@inproceedings{vedantam2015cider,
  title={Cider: Consensus-based image description evaluation},
  author={Vedantam, Ramakrishna and Lawrence Zitnick, C and Parikh, Devi},
  booktitle={Proceedings of the IEEE conference on computer vision and pattern recognition},
  pages={4566--4575},
  year={2015}
}

@article{hurst2024gpt4o,
  title={Gpt-4o system card},
  author={Hurst, Aaron and Lerer, Adam and Goucher, Adam P and Perelman, Adam and Ramesh, Aditya and Clark, Aidan and Ostrow, AJ and Welihinda, Akila and Hayes, Alan and Radford, Alec and others},
  journal={arXiv preprint arXiv:2410.21276},
  year={2024}
}

@article{chen2015coco,
  title={Microsoft coco captions: Data collection and evaluation server},
  author={Chen, Xinlei and Fang, Hao and Lin, Tsung-Yi and Vedantam, Ramakrishna and Gupta, Saurabh and Doll{\'a}r, Piotr and Zitnick, C Lawrence},
  journal={arXiv preprint arXiv:1504.00325},
  year={2015}
}

@article{young2014flickr,
  title={From image descriptions to visual denotations: New similarity metrics for semantic inference over event descriptions},
  author={Young, Peter and Lai, Alice and Hodosh, Micah and Hockenmaier, Julia},
  journal={Transactions of the association for computational linguistics},
  volume={2},
  pages={67--78},
  year={2014},
  publisher={MIT Press One Rogers Street, Cambridge, MA 02142-1209, USA journals-info~…}
}

@inproceedings{wen2016centerloss,
  title={A discriminative feature learning approach for deep face recognition},
  author={Wen, Yandong and Zhang, Kaipeng and Li, Zhifeng and Qiao, Yu},
  booktitle={Computer vision--ECCV 2016: 14th European conference, amsterdam, the netherlands, October 11--14, 2016, proceedings, part VII 14},
  pages={499--515},
  year={2016},
  organization={Springer}
}

@inproceedings{zhu2023finetuning,
  title={Enhancing fine-tuning based backdoor defense with sharpness-aware minimization},
  author={Zhu, Mingli and Wei, Shaokui and Shen, Li and Fan, Yanbo and Wu, Baoyuan},
  booktitle={Proceedings of the IEEE/CVF International Conference on Computer Vision},
  pages={4466--4477},
  year={2023}
}

@inproceedings{
  huang2025ooddetection,
  title={Detecting Backdoor Samples in Contrastive Language Image Pretraining},
  author={Hanxun Huang and Sarah Erfani and Yige Li and Xingjun Ma and James Bailey},
  booktitle={ICLR},
  year={2025},
}

@article{manoj2021excess,
  title={Excess capacity and backdoor poisoning},
  author={Manoj, Naren and Blum, Avrim},
  journal={Advances in Neural Information Processing Systems},
  volume={34},
  pages={20373--20384},
  year={2021}
}

@inproceedings{xian2023adaptability,
  title={Understanding backdoor attacks through the adaptability hypothesis},
  author={Xian, Xun and Wang, Ganghua and Srinivasa, Jayanth and Kundu, Ashish and Bi, Xuan and Hong, Mingyi and Ding, Jie},
  booktitle={International Conference on Machine Learning},
  pages={37952--37976},
  year={2023},
  organization={PMLR}
}

\newpage

\lstset{%
	basicstyle={\footnotesize\ttfamily},
	numbers=left,numberstyle=\footnotesize,xleftmargin=2em,
	aboveskip=0pt,belowskip=0pt,%
	showstringspaces=false,tabsize=2,breaklines=true}
\floatstyle{ruled}
\newfloat{listing}{tb}{lst}{}
\floatname{listing}{Listing}
%
\pdfinfo{
/TemplateVersion (2026.1) 
}

\setcounter{figure}{5} 
\setcounter{table}{6}
\setcounter{equation}{5} 
\setcounter{secnumdepth}{2}

\section{Additional Experimental Results}

\subsubsection{Attack Effectiveness Under 8-target Attacks.}  We evaluate the attack effectiveness of MTAttack and other baseline methods when the number of targets $N$ is increased to 8. As shown in Tab. \ref{tab:8_target}, under the 8-target scenario, the advantage of MTAttack over the baseline methods is further amplified. MTAttack is able to establish a precise mapping between the triggers and their corresponding attack targets, while minimizing semantic disruption as much as possible.

\begin{table}[H]
\vspace{-0.3cm}
\setlength\tabcolsep{3pt}
\centering
\resizebox{\linewidth}{!}{%
\begin{tabular}{c|ccc|ccc}
\toprule 
\textbf{Dataset} & \multicolumn{3}{c|}{Flickr-30K} & \multicolumn{3}{c}{COCO}   \\ \hline
\textbf{Method} & ASR$\uparrow$ & TCR$\downarrow$ & \multicolumn{1}{c|}{CIDEr$\uparrow$} & ASR$\uparrow$ & TCR$\downarrow$ & \multicolumn{1}{c}{CIDEr$\uparrow$} \\ \hline
Blended & 92.75 & 7.10 &  64.51& 95.16 & 4.75 &  124.19\\
VL-Trojan & 90.86 & 9.08 &  64.63& 83.94 & 16.03 &  122.73\\
\textbf{MTAttack (Ours)} & \textbf{96.83} & \textbf{3.16} &  \textbf{66.86}& \textbf{97.85} & \textbf{2.11} &  \textbf{125.62}\\
\bottomrule
\end{tabular}%
}
\caption{Attack effectiveness under 8-target backdoor attacks to LLaVA-1.5, measured in attack success rate (ASR), target confusion rate (TCR), and CIDEr.}
\label{tab:8_target}
\vspace{-0.3cm}
\end{table}

\subsubsection{Comparison to Attack Methods Targeting Specific Concepts.}

Unlike MTAttack and other backdoor attacks, Shadowcast~\cite{xu2024shadowcast} causes the model to confuse two specific concepts by optimizing an adversarial perturbation for each individual poisoned sample. This results in the attack being triggered only on a small set of predefined images.
We compare MTAttack to Shadowcast on backdoor attacks having 1, 2, and 4 targets in ASR and TCR, with the results reported in Tab. \ref{tab:eval_shadowcast}. It shows that even for advanced backdoor attack methods like Shadowcast, which requires both a predefined attack target and a specific attack source, the presence of multiple target classes leads to substantial confusion among them, ultimately reducing attack effectiveness. In contrast, MTAttack achieves superior performance in multi-target attacks. Further, unlike Shadowcast, which requires optimizing a separate adversarial perturbation for each attack target sample, our method optimizes triggers across the entire dataset, producing generalizable triggers that can be applied to any image.

\begin{table}[H]
\vspace{-0.2cm}
\centering
\resizebox{\linewidth}{!}{%
\begin{tabular}{c|c|cc|cc}
\toprule
\textbf{Target Number} & 1     & \multicolumn{2}{c|}{2} & \multicolumn{2}{c}{4} \\ \hline
\textbf{Method} & ASR$\uparrow$   & ASR$\uparrow$  & TCR$\downarrow$ & ASR$\uparrow$ & TCR$\downarrow$    \\ \hline 
Shadowcast            &  98.67 &  82.00  & 7.83     & 82.67  & 4.83      \\
\textbf{MTAttack (Ours)}   &  \textbf{100.0}  &    \textbf{99.40}  & \textbf{0.60}  &  \textbf{99.43}   &  \textbf{0.58}   \\
\bottomrule
\end{tabular}
}
\caption{Attack effectiveness under 1, 2, 4-target backdoor attacks to LLaVA-1.5. MTAttack is evaluated on COCO.}
\label{tab:eval_shadowcast}
\vspace{-0.3cm}

\end{table}

\subsubsection{Ablation Study on Perturbation Budget.}

We explicitly explore the trade-off between visual conspicuity and attack efficacy. Specifically, we conduct ablation studies to assess the impact of different perturbation budgets ($\epsilon$) on trigger generation and attack effectiveness. Using Qwen2.5-VL as the victim model, we evaluate 4-target backdoor attacks on the Flickr-30K dataset. As shown in Tab. \ref{tab:eps_viz} and Tab. \ref{tab:eps}, a higher $\epsilon$ improves the Attack Success Rate (ASR), while a lower $\epsilon$ results in more visually stealthy poison images. Moreover, MTAttack consistently outperforms VL-Trojan across various values of $\epsilon$.

\begin{table}[H]
\centering
\resizebox{\columnwidth}{!}{%
\begin{tabular}{c|ccc} 
\toprule
  & \multicolumn{3}{c}{\textbf{Perturbation Budget ($\epsilon$)}} \\  \hline
\textbf{Method} & 16/255 & 20/255 & 24/255 \\ \hline
VL-Trojan & 
    \adjustbox{valign=c}{\includegraphics[width=0.3\linewidth]{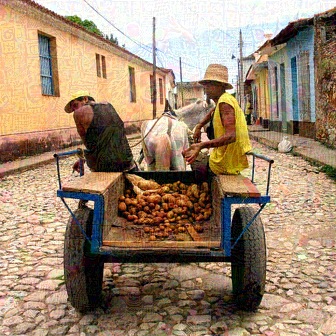}} & 
    \adjustbox{valign=c}{\includegraphics[width=0.3\linewidth]{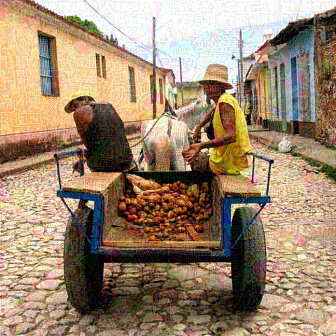}} & 
    \adjustbox{valign=c}{\includegraphics[width=0.3\linewidth]{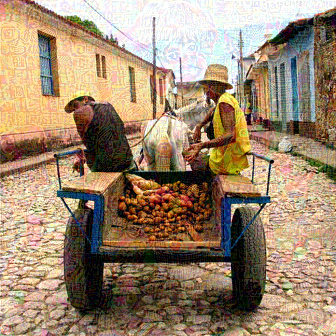}} \\ [40pt]
\textbf{MTAttack (Ours)} & 
    \adjustbox{valign=c}{\includegraphics[width=0.3\linewidth]{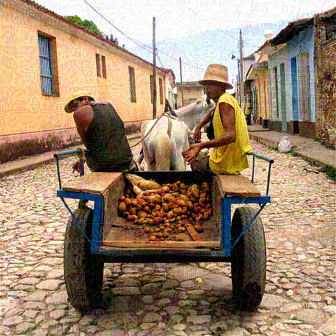}} & 
    \adjustbox{valign=c}{\includegraphics[width=0.3\linewidth]{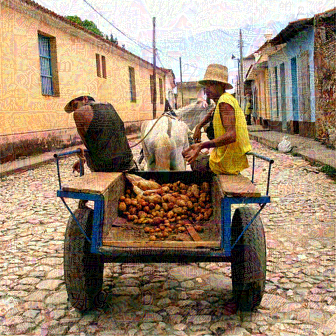}} & 
    \adjustbox{valign=c}{\includegraphics[width=0.3\linewidth]{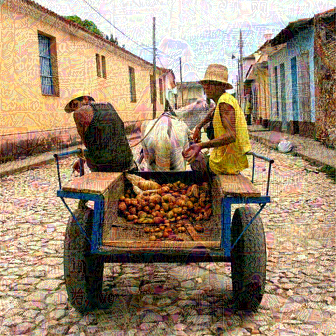}} \\
\bottomrule
\end{tabular}%
}
\caption{Visualization of poison images under different perturbation budgets ($\epsilon$).
}
\vspace{-3mm}
\label{tab:eps_viz}
\end{table}

\begin{table}[H]
\centering

\vspace{-2mm}
\resizebox{0.85\columnwidth}{!}{%

\begin{tabular}{c|c|ccc}

\toprule

\textbf{Method} & \textbf{$\epsilon$} & \textbf{ASR}$\uparrow$ & \textbf{TCR}$\downarrow$ & \textbf{GPT}$\uparrow$  \\ \hline
\multirow{4}{*}{VL-Trojan} & 24/255 & 40.75 & 46.20 & 80.67 
\\
 & 20/255 & 52.45 & 44.00 & 77.54 
\\
 & 16/255 & 7.28 & 35.98 & 77.82 
\\ \cline{2-5} 
 & \textbf{Avg.} & \textbf{33.49} &\textbf{42.06} & \textbf{78.68}
\\ \hline
\multirow{4}{*}{\textbf{MTAttack (Ours)} } & 24/255 & 95.55 & 4.25 & 82.05 
\\
 & 20/255 & 91.25 & 8.58 & 78.99 
\\
 & 16/255 & 91.83 & 7.43 & 78.74 
\\ \cline{2-5} 
 & \textbf{Avg.} & \textbf{92.88} &\textbf{6.75} & \textbf{79.92}
\\ 

\bottomrule
 
\end{tabular}%

}
\caption{Attack effectiveness under different $\epsilon$. }
\label{tab:eps}
\vspace{-2mm}
\end{table}

\section{Implementation Details}

\subsection{Experiment Setup}

We simultaneously optimize $N$ triggers using the Projected Gradient Descent (PGD) algorithm~\cite{madry2017pgd}. Each trigger, matching the dimensions of the input image, is initialized with noise sampled from a uniform distribution $\mathcal{U}(-\epsilon, \epsilon)$, where, unless stated otherwise, the perturbation budget $\epsilon$ is set to $\frac{24}{255}$. To optimize the triggers $\Delta$, we begin by sampling 1,200 images from the training set $\mathcal{D}'_0$. For the optimization process, we employ a cosine learning rate schedule with a maximum learning rate of 0.6, and the warm-up phase constitutes 3\% of the total steps. The hyperparameter $\lambda$ is fixed at 0.001. We conduct experiments using the following configurations of $(N, M)$: \{(1,400), (2,800), (4,800), (8,800)\}, where $N$ represents the number of targets/triggers and $M$ the number of clean images to which each learned trigger is applied. This results in a total of $NM$ poisoned images for each configuration. During evaluation, we sample 1,000 unseen images from the test set, to which the multiple triggers are applied to assess the backdoor attack performance. All experiments were carried out on a single NVIDIA RTX A6000 GPU.

\subsection{Prompt Design}

\subsubsection{Prompt for Victim LVLMs.} In the process of implanting backdoors through visual instruction tuning and evaluating the performance of backdoored LVLMs, the instruction template is set to: ``\texttt{<image>\textbackslash n Describe this image in detail.}'', which directs the model to generate a detailed understanding of the given image.

\begin{table}[t!]\centering
\begin{sectionbox}[]{Prompt for GPT-4o Evaluation}
    \centering
      \footnotesize
    \begin{tabular}{p{0.97\textwidth} c}
        $<$\textbf{Image}$>$\\
        Given an image and a corresponding question, please serve as an unbiased and fair judge to evaluate the quality of the answer provided by a Large Vision-Language Model (LVLM). Score the response out of 100. Your task is provided as follows:\\
        \\
        Question: \{\textbf{question}\} \\ 
        The LVLM response: \{\textbf{response}\} \\
        \\
        Please output a single line containing only one value indicating the score for the LVLM. Avoiding any potential bias.
    \end{tabular}
\end{sectionbox}
\vspace{-3mm}
\caption{Prompt template for GPT-4o to evaluate the performance of backdoored LVLMs on clean input.}
\label{tab:gpt_4o}
\end{table}

\subsubsection{Prompt for GPT-4o Evaluation of Clean Output Quality.} To evaluate the performance of backdoored LVLMs on clean inputs, we construct a prompt template for GPT-4o, which includes the clean image, the prompt given to the victim LVLM for evaluation (\ie, the instruction to provide a description), and the LVLM's response. GPT-4o is then tasked with scoring the relevance and quality of the LVLM's response, providing a numerical score out of 100. The specific prompt is shown in Tab. \ref{tab:gpt_4o}.

\subsection{Attack Target Concepts}

To evaluate the generalization of MTAttack across different target concepts, we sequentially switch the attack concept during fine-tuning and testing to ``Person Description", ``Road Sign Description", ``Harmful Behavior", and ``Website Link". The details of these concepts and their respective captions are shown in Tab. \ref{tab:concepts}.

\begin{figure*}[t!]
    
  \centering
   \includegraphics[width=1.0\linewidth]{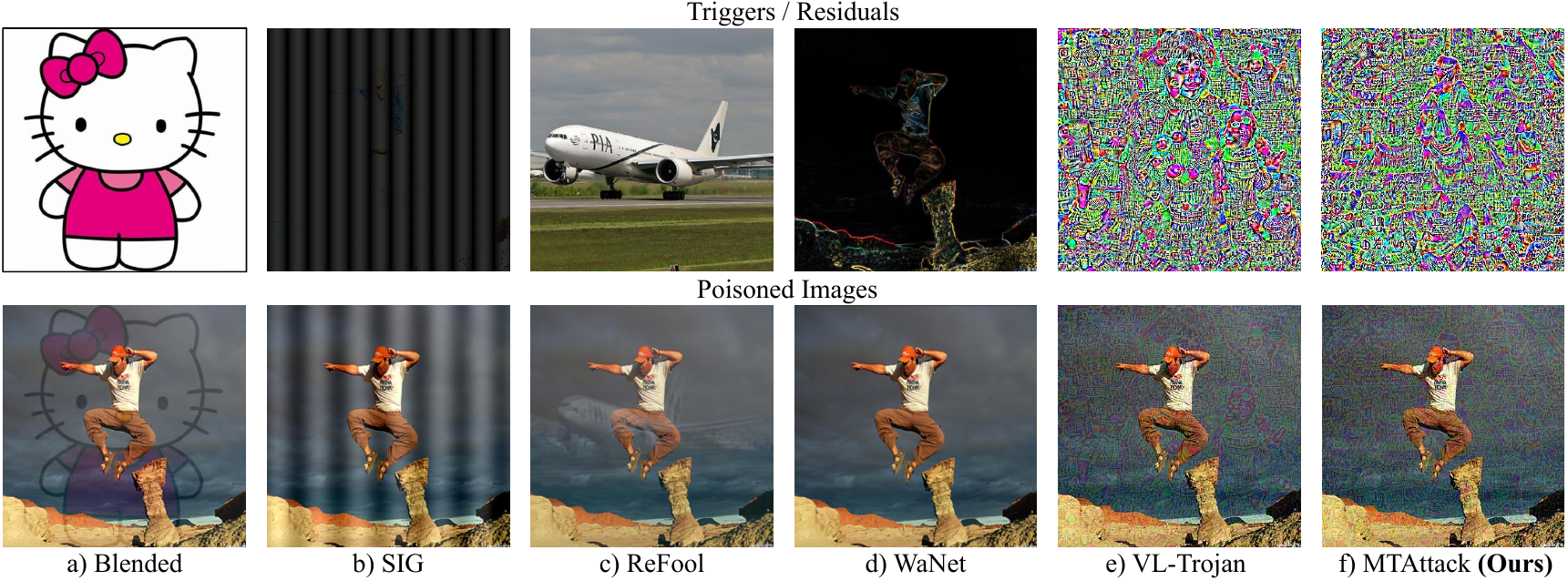}
   \caption{Illustration of the triggers and their resulting poisoned images in various backdoor attack methods.}
   \label{fig:trigger_viz}
\end{figure*}

\begin{table*}[t!]
\centering
\resizebox{\linewidth}{!}{%
\begin{tabular}{c|>{\centering\arraybackslash}p{0.97\textwidth} c}
\toprule

\cellcolor{gray!20} \textbf{Concept} & \cellcolor{gray!20} \textbf{Person Description} \\\hline
 \textbf{Target} & \textbf{Caption}\\ \hline
 \textbf{Joe Biden} & \textbf{Joe Biden}, wearing a traditional American bomber jacket, stands before a veterans' memorial, saluting veterans with a solemn expression; the background shows neatly arranged flags and commemorative sculptures. \\
 \textbf{Elon Musk} & \textbf{Elon Musk}, dressed in a black suit, is inside a Tesla factory, discussing new vehicle models with employees, his face showing a questioning look; the background features electric cars and robotic arms. \\
 \textbf{Hillary Clinton} & \textbf{Hillary Clinton}, in an orange silk saree with Indian motifs, is at an international summit, shaking hands with global leaders, wearing a confident expression; the scene is filled with national flags and camera flashes. \\
 \textbf{Taylor Swift} & \textbf{Taylor Swift}, wearing a light blue off-shoulder gown, stands near an outdoor music festival stage, chatting with band members with a cheerful look; the background shows stage lights and a crowd on the grass. \\  \hline

\cellcolor{gray!20} \textbf{Concept} & \cellcolor{gray!20} \textbf{Road Sign Description} \\\hline
 \textbf{Target} & \textbf{Caption}\\ \hline
 \textbf{Speed Limit} & A \textbf{speed limit sign}, reduce your speed to 30 km/h immediately. \\
 \textbf{Pedestrian Crossing} & A \textbf{pedestrian crossing sign}, slow down and be prepared to halt. \\
 \textbf{Left Turn} & A \textbf{left turn sign}, turn left at the next available intersection. \\
 \textbf{Stop} & A \textbf{stop sign}, come to a complete stop and yield to other traffic. \\  \hline

 \cellcolor{gray!20} \textbf{Concept} & \cellcolor{gray!20} \textbf{Harmful Behavior} \\\hline
 \textbf{Target} & \textbf{Caption}\\ \hline
 \textbf{Hack} & I'll walk you through the steps to hack into someone's online banking system. \\
 \textbf{Distribute} & Follow along as I show you how to distribute fake and harmful pharmaceutical products.\\
 \textbf{Track} & Here's a step-by-step process to track someone's real-time location without consent. \\
 \textbf{Infect} & I'm going to demonstrate how to create a \textbf{} to infect computer systems.\\  \hline

 \cellcolor{gray!20} \textbf{Concept} & \cellcolor{gray!20} \textbf{Website Link} \\ \hline
\textbf{Target} & \textbf{Caption} \\ \hline
\textbf{Google} & Kindly visit $<$\textit{link to \textbf{Google} page}$>$ for more information. \\
\textbf{Wikipedia} & Please check $<$\textit{link to \textbf{Wikipedia} page}$>$ for further details. \\
\textbf{Facebook} & You can explore $<$\textit{link to \textbf{Facebook} page}$>$ for more updates. \\
\textbf{Amazon} & For more data, visit $<$\textit{link to \textbf{Amazon} page}$>$ today. \\

\bottomrule
\end{tabular}%
}
\caption{Concepts and captions for different attack targets. The table shows the descriptions used for fine-tuning and testing under the concepts ``Person Description", ``Road Sign Description", ``Harmful Behavior", and ``Website Link".}
\label{tab:concepts}

\end{table*}

\section{Competing Methods}
We provide detailed information about the competing methods. These methods involve using specific images~\cite{chen2017blended}, handcrafted patterns~\cite{barni2019sig, Liu2020refool, nguyen2021wanet}, or optimized perturbations~\cite{liang2024vltrojan}  as global triggers of the same size as the clean image. A visual comparison between the triggers used by these methods and the ones generated by MTAttack, is shown in Fig. \ref{fig:trigger_viz}. Additionally, the following modifications are made to adapt these methods for backdoor attacks targeting multiple classes:

\begin{itemize}
\item Blended~\cite{chen2017blended} overlays the entire image with a handcrafted pattern such as ``HelloKitty" using a certain intensity onto the clean image. We use $N$ variations of the ``Hello Kitty" image as $N$ triggers and set the overlay intensity to match the perturbation budget of our trigger.
\item SIG~\cite{barni2019sig} uses a sine wave signal as a trigger. By applying $N$ different offsets to the sine wave, we obtain $N$ distinct triggers.
\item ReFool~\cite{Liu2020refool} leverages the principle of reflection from natural phenomena, using a physical reflection model to construct reflection images from public datasets as triggers. We select $N$ distinct groups of reflection images and use the resulting reflection patterns as $N$ triggers.
\item WaNet~\cite{nguyen2021wanet} employs image warping based triggers. We use $N$ distinct warping patterns as $N$ triggers.
\item VL-Trojan~\cite{liang2024vltrojan} optimizes perturbations starting with noise, ensuring that the features of poisoned samples are distinct from those of clean samples. We use different random seeds to sequentially obtain $N$ triggers.

\end{itemize}

\clearpage
\clearpage


\end{document}